\documentclass{article}

\PassOptionsToPackage{numbers, sort}{natbib}

\usepackage{xspace}

\makeatletter
\DeclareRobustCommand\onedot{\futurelet\@let@token\@onedot}
\def\@onedot{\ifx\@let@token.\else.\null\fi\xspace}

\def\eg{\emph{e.g}\onedot}

\def\etc{\emph{etc}\onedot} \def\vs{\emph{vs}\onedot}
 
\def\etal{\emph{et al}\onedot}
\makeatother

\usepackage[preprint]{neurips_2025}

\usepackage[utf8]{inputenc} 
\usepackage[T1]{fontenc}    
\usepackage{hyperref}       
\usepackage{url}            
\usepackage{booktabs}       
\usepackage{amsfonts}       
\usepackage{nicefrac}       
\usepackage{microtype}      
\usepackage[numbers,sort]{natbib}
\usepackage{amsmath, amssymb}

\usepackage{pifont}
\usepackage{booktabs}
\usepackage{adjustbox}
\usepackage{bbding}
\usepackage{graphicx}
\usepackage{multirow}
\usepackage{enumitem}
\usepackage{subcaption}
\usepackage[table,xcdraw]{xcolor}
\usepackage[most]{tcolorbox}

\definecolor{lightyellow}{RGB}{255, 255, 204}
\definecolor{beaublue}{RGB}{240, 247, 255}
\hypersetup{
    colorlinks=true,
    citecolor=blue, 
}

\title{VAU-R1: Advancing Video Anomaly Understanding via Reinforcement Fine-Tuning}

\author{%
  Liyun Zhu$^{1,2,}$\thanks{Work done while the author was a visiting student at GVC Lab, Great Bay University.} \quad Qixiang Chen$^{1}$ \quad Xi Shen$^{3}$ \quad Xiaodong Cun$^{2,}$\thanks{Corresponding Author} \\
  \textsuperscript{\rm 1}~Australian National University \quad \textsuperscript{\rm 2}~GVC Lab, Great Bay University\\ 
  \textsuperscript{\rm 3}~Intellindust AI Lab\\
  \texttt{\{liyun.zhu, u7227010\}@anu.edu.au,} \\
  \texttt{shenxiluc@gmail.com, cun@gbu.edu.cn}
}

\begin{document}

\maketitle

\vspace{-0.3cm}
\begin{abstract}
\vspace{-0.1cm}
Video Anomaly Understanding (VAU) is essential for applications such as smart cities, security surveillance, and disaster alert systems, yet remains challenging due to its demand for fine-grained spatio-temporal perception and robust reasoning under ambiguity. Despite advances in anomaly detection, existing methods often lack interpretability and struggle to capture the causal and contextual aspects of abnormal events. This limitation is further compounded by the absence of comprehensive benchmarks for evaluating reasoning ability in anomaly scenarios. To address both challenges, we introduce \textbf{VAU-R1}, a data-efficient framework built upon Multimodal Large Language Models (MLLMs), which enhances anomaly reasoning through Reinforcement Fine-Tuning (RFT). Besides, we propose \textbf{VAU-Bench}, the first Chain-of-Thought benchmark tailored for video anomaly reasoning, featuring multiple-choice QA, detailed rationales, temporal annotations, and descriptive captions. Empirical results show that VAU-R1 significantly improves question answering accuracy, temporal grounding, and reasoning coherence across diverse contexts. Together, our method and benchmark establish a strong foundation for interpretable and reasoning-aware video anomaly understanding. Our code is available at
\href{https://github.com/GVCLab/VAU-R1}{\texttt{https://github.com/GVCLab/VAU-R1}}.
\end{abstract}

\begin{figure}[h]
    \centering
    \vspace{-1em}
    \includegraphics[width=0.85\linewidth, trim=5 7 5 12, clip=true]{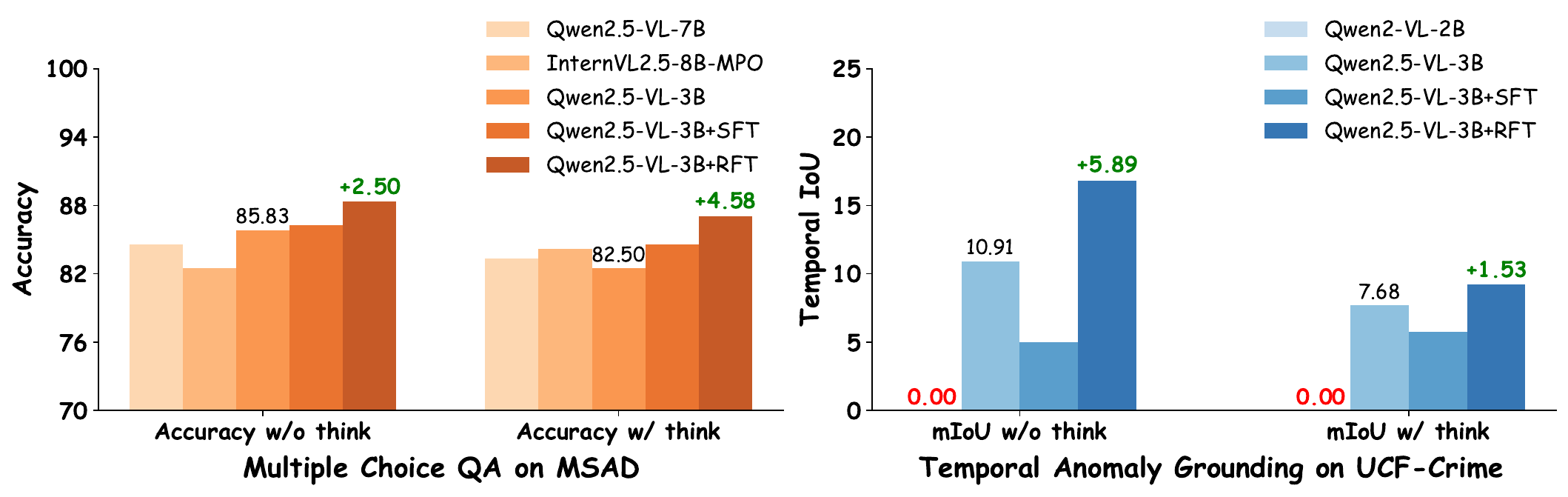}
    \caption{\textbf{Effectiveness of Reinforcement Fine-Tuning.} We compare QA accuracy and temporal anomaly grounding performance across different models. VAU-R1, trained via Reinforcement Fine-Tuning~(RFT), consistently outperforms its Supervised Fine-Tuning~(SFT) counterpart. This demonstrates that RFT enhances both reasoning and temporal localization capabilities in VAU tasks. }
    \label{fig:performance}
\end{figure}

\section{Introduction}



\begin{figure}[ht]
    \centering
    \includegraphics[width=\linewidth]{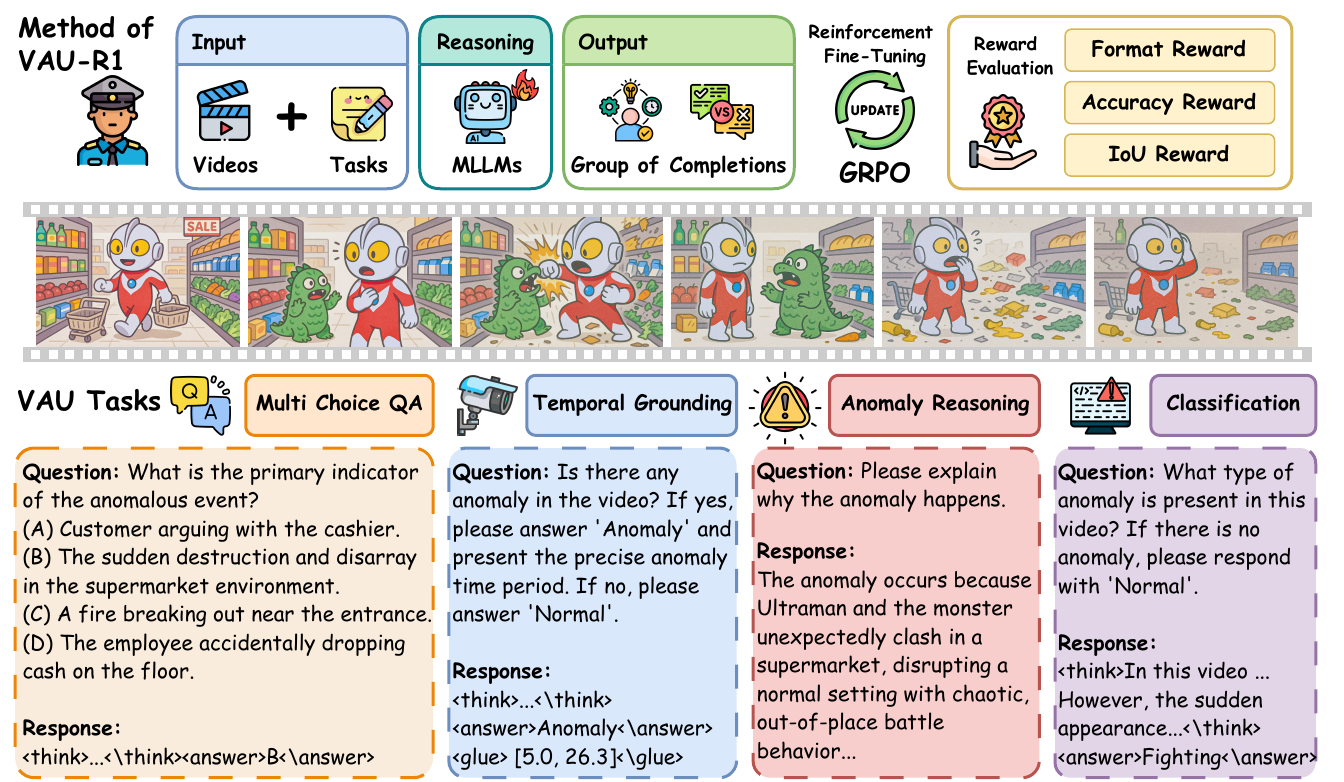}
    \caption{\textbf{Overview of VAU-R1.} VAU-R1 leverages Reinforcement Fine-Tuning to enhance the reasoning ability of MLLMs for video anomaly understanding. Specifically, we adopt Group Relative Policy Optimization (GRPO) to optimize the model with task-specific rewards, such as answer format, accuracy, and temporal Intersection-over-Union~(IoU). We decompose the VAU task into four complementary tasks to facilitate comprehensive reasoning: multiple-choice QA, temporal anomaly grounding, anomaly reasoning, and anomaly classification.}
    \vspace{-0.5cm}
    \label{fig:overview}
\end{figure}

Anomalies are events or behaviors that deviate from regular patterns or expected activities in a given context. In surveillance settings, these may include incidents such as fighting, theft, or traffic violations, \etc. Video Anomaly Understanding (VAU) aims to detect and interpret such irregular events in unstructured, real-world video streams~\cite{liu2024generalized}. The task is challenging due to scene complexity, context dependence, varying camera viewpoints, and diverse anomaly types~\cite{wu2024deep,pang2021deep,zhuadvancing}. Early approaches only focuses on detecting anomalies, which typically framed the task as binary classification, assigning normal or abnormal labels to individual frames and identifying the temporal boundaries of anomalous events~\cite{gong2019memorizing,liu2018future,tian2021weakly,wu2024vadclip,lengbeyond,zhou2023dual,sultani2018real,chen2023mgfn,ding2024lego}. While effective for localization, these methods offer limited interpretability and provide little insight into the underlying causes of anomalies~\cite{ye2024vera,doshi2023towards,du2024exploring}. Recent advances in Multi-modal Large Language Models (MLLMs) have introduced the ability to generate textual descriptions of anomalous events~\cite{yang2024follow,zanella2024harnessing,du2024uncovering,zhang2024holmes,zhang2024holmesvad}, improving model transparency to some extent. However, current methods still face three key limitations: (i) they lack the ability to generate coherent, multi-step reasoning chains; (ii) no comprehensive benchmark provides rich annotations to support detailed causal reasoning; and (iii) evaluation protocols for reasoning quality remain underdeveloped.

To move beyond shallow classification and toward deeper understanding, we decompose VAU into four progressive stages: (i)~\textbf{\textit{Perception}} — identifying the scene and relevant objects, either through free-text descriptions or guided multiple-choice questions; (ii)~\textbf{\textit{Grounding}} — localizing the precise temporal segment where the anomaly occurs; (iii)~\textbf{\textit{Reasoning}} — explaining the event by analyzing causal factors, temporal dynamics, and contextual cues; and (iv)~\textbf{\textit{Conclusion}} — summarizing the event with a final decision, such as assigning it to a specific category (e.g., \textit{fighting} vs.\ \textit{robbery}). This structured formulation enables models to progressively build semantic understanding and supports more interpretable and task-aligned evaluation.

To implement this four-stage formulation, we introduce \textbf{VAU-R1}, a Reinforcement Fine-Tuning (RFT) framework designed to improve the reasoning capabilities of MLLMs on the VAU task. Our method builds on Group Relative Policy Optimization  (GRPO)~\cite{shao2024deepseekmath}, incorporating task-specific reward signals based on answer format correctness, question-answer accuracy, and temporal grounding alignment. The framework is data-efficient and can be applied in low-resource settings, making it practical for real-world deployments. To support training and evaluation, we also construct \textbf{VAU-Bench}, a new benchmark that spans diverse scenarios and provides rich annotations across the four reasoning stages, including multiple-choice QA pairs, detailed event descriptions, temporal groundings, and step-by-step rationales. Finally, we propose a set of evaluation metrics—QA accuracy, temporal Intersection-over-Union~(IoU), GPT-based reasoning score, and classification accuracy—to quantitatively assess model performance across perception, grounding, reasoning, and conclusion. Together, VAU-R1 and VAU-Bench offer a scalable and unified framework for advancing structured video anomaly understanding. Our contribution can be summarized as follows:


\begin{itemize}
    \setlength\itemsep{-0.2em}  
  \setlength\topsep{-0.3cm}  
    \item We propose \textbf{VAU-R1}, a data-efficient Reinforcement Fine-Tuning framework that improves the reasoning ability of MLLMs for video anomaly understanding. It outperforms standard supervised fine-tuning on reasoning-intensive tasks.

    \item We present \textbf{VAU-Bench}, the first large-scale benchmark with Chain-of-Thought annotations designed for video anomaly reasoning. It contains a diverse collection of videos, QA pairs, temporal labels, and detailed rationales spanning a wide range of real-world scenarios.

    \item We design a unified evaluation protocol that measures model performance across four reasoning stages, jointly considering reasoning quality, answer correctness, and temporal localization to capture both interpretability and detection precision.
\vspace{-0.2cm}
\end{itemize}

\section{Related Works}

\noindent{\textbf{From Detection to Understanding.}} Early efforts in Video Anomaly Detection (VAD) can be broadly categorized into self-supervised and weakly-supervised paradigms. Self-supervised methods rely solely on normal video samples, learning the distribution of normal behavior and flagging deviations as anomalies~\cite{liu2018future,gong2019memorizing,lu2020few}. In contrast, weakly-supervised methods are trained with both normal and anomalous videos using coarse video-level labels rather than fine-grained frame-level annotations~\cite{sultani2018real,tian2021weakly,wu2024vadclip,lengbeyond,zhou2023dual,chen2023mgfn,wu2024open}. These approaches typically adopt a \textit{top-$k$} selection strategy to identify the most likely anomalous segments. While effective for localizing anomaly boundaries, they often rely heavily on motion cues~\cite{zhuadvancing}, operating under the assumption that rapid or irregular motion is indicative of anomalous behavior. However, this assumption does not hold for subtle or semantically complex anomalies, leading to poor interpretability. To address these limitations, recent work has turned to video anomaly understanding, leveraging MLLMs to provide more semantically grounded and interpretable reasoning~\cite{lv2024video}.

\noindent{\textbf{Prompt-Based \vs Learning-Based Approaches for VAU.}} Building on the shift toward semantic understanding, recent approaches to VAU fall into two main categories: prompt-based and learning-based methods. Prompt-based methods typically use MLLMs as anomaly scorers~\cite{zanella2024harnessing,shao2025eventvad}, or as reasoning agents via rule-based few-shot prompting~\cite{yang2024follow} or learned question templates~\cite{ye2024vera}. While these methods avoid computationally expensive training, their generalization ability is often limited due to the absence of task-specific adaptation. On the other hand, pretraining~\cite{du2024exploring} and finetuning~\cite{zhang2024holmes,zhang2024holmesvad} approaches aim to learn anomaly-aware representations by incorporating video captions and causal reasoning signals (\eg, cause and effect). Despite this progress, existing methods remain constrained to improving anomaly description and fail to capture the full logical chain of an anomaly. To overcome these limitations, we leverage reinforcement fine-tuning to enhance the model’s reasoning ability, enabling end-to-end identification of both when and why anomalies occur.

\noindent{\textbf{Reinforcement Learning in MLLMs.}}
With the rise of powerful models such as OpenAI-o1~\cite{jaech2024openai} and DeepSeek-R1~\cite{guo2025deepseek}, reinforcement learning has been increasingly adopted in the post-training stage of MLLMs to enhance their reasoning capabilities~\cite{feng2025video,wang2024enhancing,huang2025vision,zhou2025r1,bi2025reasoning}. While effective, this process often demands substantial computational resources and large-scale datasets, making it less practical for targeted downstream tasks~\cite{tan2025reason}. To address these challenges, Visual-RFT~\cite{liu2025visual} introduces Reinforcement Fine-Tuning~(RFT) for visual tasks, demonstrating improved data efficiency and stronger performance compared to Supervised Fine-Tuning~(SFT). Building on this idea, VideoChat-R1~\cite{li2025videochat} extends RFT to video domains, achieving promising results in tasks such as question answering, temporal grounding, and object tracking. Yet, these tasks remain fragmented and have not been unified under the video anomaly understanding setting. To bridge this gap, we propose a framework that jointly addresses multiple tasks, aiming to advance comprehensive and interpretable anomaly reasoning.

\section{Methodology}

\subsection{Preliminary: Reinforcement Learning via Group Relative Policy Optimization}
Group Relative Policy Optimization (GRPO)~\cite{shao2024deepseekmath} is a reinforcement learning framework that optimizes a policy~$\pi_\theta$ using preference-based feedback and multi-aspect reward signals. Given a question~$x$, GRPO generates $M$ candidate outputs $O = \{o_1, o_2, \ldots, o_M\}$  from the old policy $\pi_{\theta_{\text{old}}}$, each output $o_j$ assigned a reward $R_j$ computed as a weighted sum of $K$ task-specific components:
\begin{equation}
R_j = \sum_{k=1}^{K} \lambda_k R_j^{(k)},
\end{equation}
where $R^{(k)}_j$ is the $k$-th task-specific reward (\eg, accuracy, IoU, format compliance) and
$\lambda_k$ is its weight. To measure the relative quality of the $j$-th output, we calculate the normalized reward $\tilde{R}_j$ for each output $o_j$ with the mean $\mu_R$ and standard deviation $\sigma_R$ across $M$ candidates:\
\begin{equation}
\tilde{R}_j = \frac{R_j - \mu_R}{\sigma_R}.
\end{equation}
GRPO maximises the following objective while keeping the update close to the original MLLM parameters~$\pi_{\text{ref}}$ through a KL penalty term $\mathrm{D}_{\mathrm{KL}}(\cdot \mid\mid \cdot)$:

\begin{equation}
\max_{\pi_\theta} \mathbb{E}_{O \sim \pi_{\theta_{\text{old}}}(\cdot\mid x)} \left[ 
\sum_{j=1}^{M} \frac{\pi_\theta(o_j)}{\pi_{\theta_{\text{old}}}(o_j)} \cdot \tilde{R}_j 
- \beta \, \mathrm{D}_{\mathrm{KL}}\left( \pi_\theta \,\|\, \pi_{\text{ref}} \right) 
\right],
\end{equation}
where $\beta$ is a regularization coefficient. This formulation allows GRPO to incorporate diverse reward signals while retaining training stability through KL regularization.

\subsection{VAU-R1}


As shown in Figure~\ref{fig:overview}, VAU-R1 is a data-efficient reinforcement fine-tuning framework designed for the four VAU tasks, including Multi-choice QA, Temporal Anomaly Grounding, Anomaly Reasoning, and Anomaly Classification. Given videos and task-specific questions, we fine-tune a pre-trained MLLM to improve its multi-step reasoning ability across different tasks. The model generates multiple candidate responses for each input, which are then scored using task-specific reward functions (\eg, accuracy, temporal IoU, or format compliance). We employ Group Relative Policy Optimization~(GRPO) to optimize the model, which maximizes reward-weighted likelihood while constraining divergence from the reference model via KL regularization. Our reinforcement-based approach outperforms supervised fine-tuning~(SFT) in both reasoning capability and generalization to unseen scenarios. The design of task-specific reward functions is further detailed in Section~\ref{sec:reward}.



\subsection{Reward Rules}
\label{sec:reward}
We adopt the general idea of GRPO-based RFT to optimize the VAU model by designing task-specific reward functions for different VAU components. Below, we detail each reward definition.


\noindent\textbf{Format Reward.} For multiple-choice QA and anomaly classification tasks, we instruct the model to enclose its reasoning within \texttt{<think>...</think>} tags and the answer within \texttt{<answer>...</answer>} tags.  
For the temporal anomaly grounding task, we additionally require \texttt{<glue>...</glue>} tags to indicate the predicted time span in seconds. The reward is defined as:
\begin{equation}
  R_{\text{format}} = 
  \begin{cases}
    1, & \text{if the output format is correct},\\
    0, & \text{otherwise}.
  \end{cases}
\end{equation}
We apply a format reward to VAU tasks to enforce structured outputs and discourage format violations.

\noindent\textbf{Accuracy Reward.}  
We also define an accuracy reward $R_{\text{acc}}$ to measure the correctness of the model’s answer. In our experiments, this reward is given by:
\begin{equation}
  R_{\text{acc}} = 
  \begin{cases}
    1, & \text{if output = ground truth},\\
    0, & \text{otherwise}.
  \end{cases}
\end{equation}
This simple accuracy reward encourages the model to choose the right answer during training.  

\noindent\textbf{Temporal IoU Reward.}  
To encourage precise temporal grounding, we introduce a temporal Intersection-over-Union (IoU) reward $R_{\text{tIoU}}$, which measures the alignment between the predicted and ground truth anomaly intervals.  
The reward is defined as:
\begin{equation}
  R_{\text{tIoU}} =
  \begin{cases}
    1, & \text{if the model correctly classifies a normal video};\\
    \mathrm{IoU}([s_1, s_2], [s^*_1, s^*_2]), & \text{if the model correctly detects an anomaly and predicts } [s_1, s_2];\\
    0, & \text{otherwise}.
  \end{cases}
\end{equation}
Here, \([s_1, s_2]\) denotes the predicted temporal span of the anomaly, while \([s^*_1, s^*_2]\) is the ground truth interval.  
The temporal IoU quantifies the degree of overlap between these intervals, and serves as a fine-grained reward signal to guide the model toward more accurate temporal localization.

\begin{figure}[t]
    \centering
    \begin{subfigure}[ht]{0.45\linewidth}
        \centering
        \includegraphics[width=\linewidth, trim=40 50 40 40, clip=true]{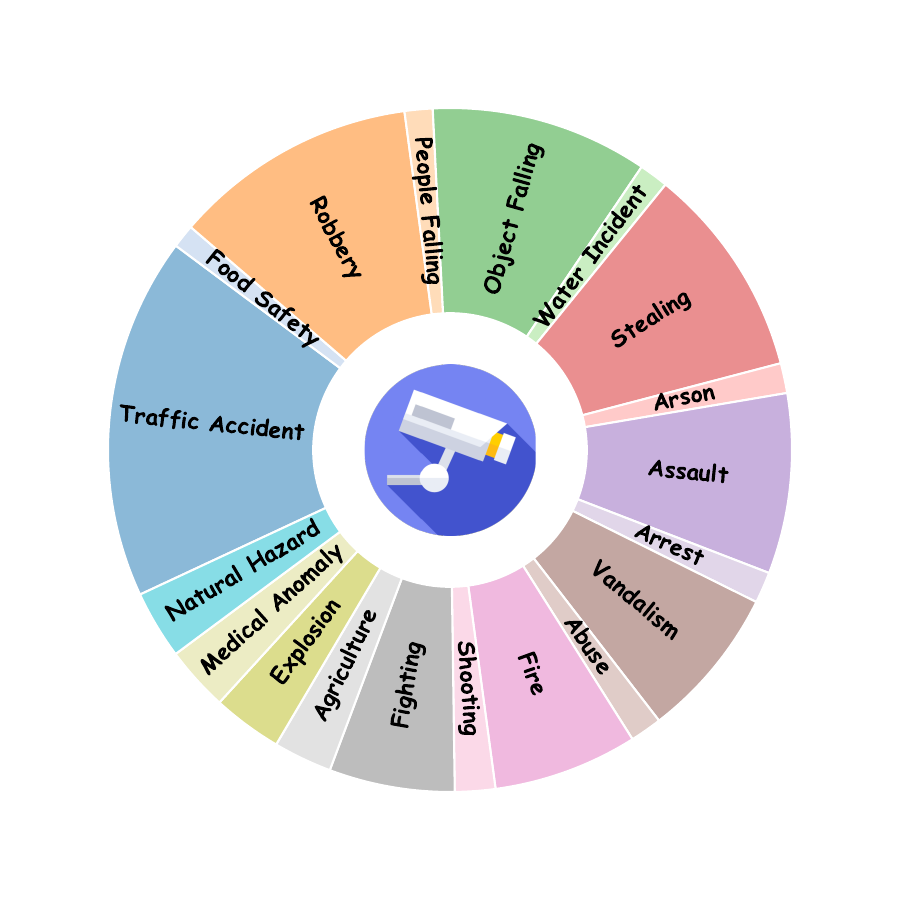}
        \caption{}
        \label{fig:anomaly-type}
    \end{subfigure}
    \hspace{0.5cm}
    \begin{subfigure}[ht]{0.425\linewidth}
        \centering
        \includegraphics[width=\linewidth,trim=-8 12 -5 -25, clip=true]{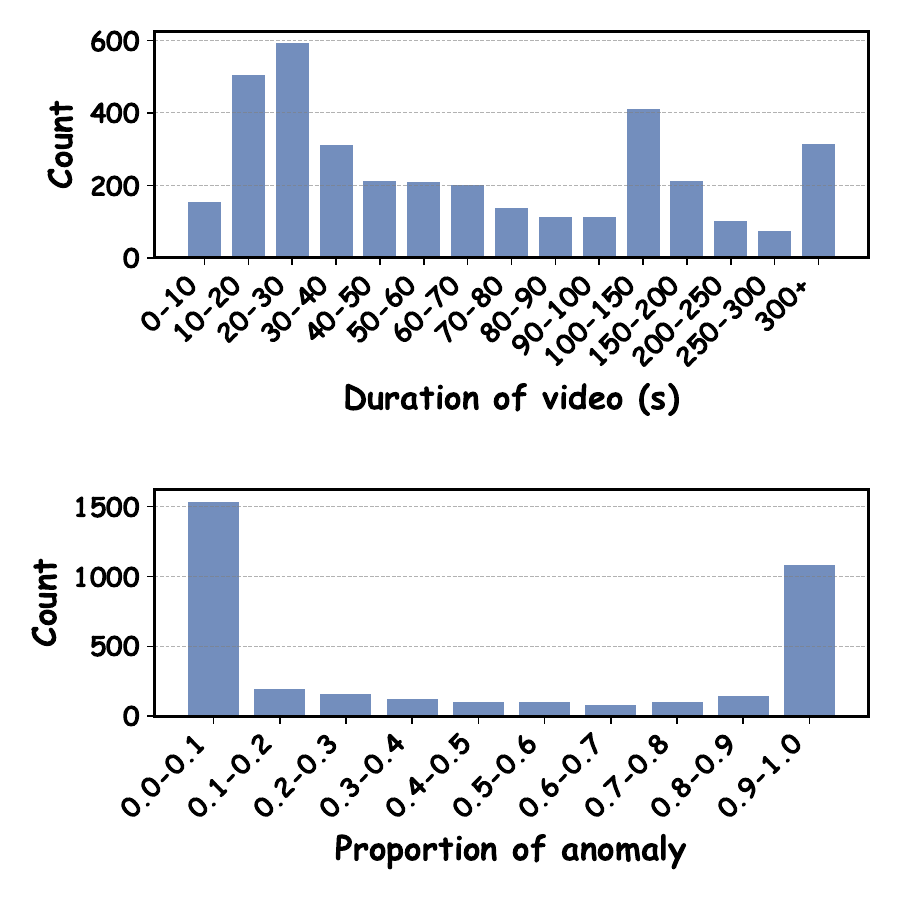}\\[1ex]
        \caption{}
        \label{fig:video-length-prop}
    \end{subfigure}
    \begin{subfigure}[ht]{\linewidth}
        \centering
        \includegraphics[width=\linewidth,trim=0 5 0 -5, clip=true]{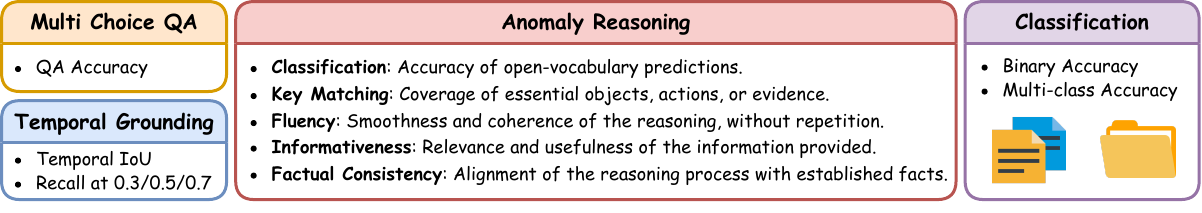}
        \caption{}
        \label{fig:eval-criteria}
    \end{subfigure}
    \caption{\textbf{Statistics of our VAU‐Bench.} 
    (a) Distribution of main anomaly types.
    (b) Distribution of video durations \textit{(top)} and the proportion of anomalous segments within each video \textit{(bottom)}.  
    (c) The evaluation criteria for four VAU tasks.}
    \label{fig:dataset-stat}
    \vspace{-0.3cm}
\end{figure}

\noindent\textbf{Task-specific Reward Formulations.}  
We apply task-specific combinations of the reward components mentioned above. For the multiple-choice QA task, we use a combination of format and accuracy rewards: $R_{\text{QA}} = R_{\text{format}} + R_{\text{acc}}$. For temporal anomaly grounding, we further include a temporal IoU term to evaluate localization quality:  
$R_{\text{TAG}} = R_{\text{format}} + R_{\text{acc}} + R_{\text{tIoU}}$.  
For anomaly classification, we adopt a similar reward design as QA:  
$R_{\text{CLS}} = R_{\text{format}} + R_{\text{acc}}$.

\subsection{VAU-Bench}
\label{sec:benchmark}

\noindent\textbf{Task Definition.} We decompose the VAU task into four stages: perception, grounding, reasoning, and conclusion. These stages address four core questions respectively: "\textit{What happens in this video?}", "\textit{When does the anomaly occur?}", "\textit{Why does the anomaly happen?}", and "\textit{What is our overall judgment of the anomaly?}". Corresponding to these stages, we define four VAU tasks:

\begin{itemize}[leftmargin=20pt, topsep=0pt]
  \item \textbf{Multiple-Choice QA}: Targets event perception by answering questions about videos.
  \item \textbf{Temporal Grounding}: Localizes anomalous segments in the video timeline.
  \item \textbf{Anomaly Reasoning}: Explores causal relationships to explain why an anomaly arises.
  \item \textbf{Anomaly Classification}: Assigns the anomaly to its corresponding category.
\end{itemize}
 
This structured decomposition provides a clear framework for systematically addressing different perspectives of VAU, with each task rigorously evaluated using domain-specific metrics.

\begin{table*}[t]
    \centering
    \setlength{\tabcolsep}{3pt}
    \caption{
    Comparison of performance on MSAD and UCF-Crime datasets on multiple-choice QA task and anomaly reasoning task.
    \textbf{Acc}$_{\text{w/o think}}$ and \textbf{Acc}$_{\text{w/ think}}$ refer to the multiple-choice question accuracy without and with thinking, respectively. For the anomaly reasoning task,
    \textbf{CLS}, \textbf{KM}, \textbf{FLU}, \textbf{INF}, and \textbf{FAC} represent VAU-Eval scores generated by DeepSeek-V3, measuring classification accuracy, key concept alignment, linguistic fluency, informativeness, and factual consistency, respectively. Each dimension is scored on a 10-point scale. \textbf{Total} denotes the aggregated score over five dimensions.
    }

    \resizebox{\textwidth}{!}{\begin{tabular}{llllllllll}
    \toprule
    \multirow{2.5}{*}{\textbf{Dataset}} & \multirow{2.5}{*}{\textbf{Model}} & \multicolumn{2}{c}{\textbf{QA Accuracy}} & \multicolumn{6}{c}{\textbf{VAU-Eval}} \\ 
    \cmidrule(lr){3-4} \cmidrule(lr){5-10}&     & \textbf{Acc}$_{\text{w/o think}}$ & \textbf{Acc}$_{\text{w/ think}}$ & \textbf{CLS}$\uparrow$ & \textbf{KM}$\uparrow$ & \textbf{FLU}$\uparrow$ & \textbf{INF}$\uparrow$ & \textbf{FAC}$\uparrow$ & \textbf{Total}$\uparrow$ \\ 
    \midrule
    \multirow{10}{*}{MSAD} & InternVL2.5-2B & 76.67 & 72.08 & 6.84 & 6.23 & 8.55 & 6.64 & 6.64 & 34.90 \\
    & Qwen2.5-VL-7B & 84.58 & 83.33 & 6.75 & 6.41 & 9.27 & 7.74 & 6.92 & 37.08\\
    & InternVL2.5-8B-MPO & 82.50 & 84.17   & 6.83 & 6.33 & 8.32 & 6.37 & 6.86 & 34.72\\
    \cmidrule(lr){2-10}
    & Qwen2-VL-2B & 77.08 & 72.50 & 5.94 & 5.43 & 8.77 & 6.29 & 5.90 & 32.25\\
    & ~+\textcolor{teal!90}{SFT}  &  82.92  & \textbf{85.83} & 6.04 & 5.43 & 8.89 & 6.55 & 5.93 & 32.84 \\
    & ~+\textcolor{red!85}{RFT}  & \textbf{82.92~(\textcolor{red}{$\uparrow$5.84})} & 83.75~\textbf{(\textcolor{red}{$\uparrow$11.25})} & 6.05(\textcolor{red}{$\uparrow$}) & 5.49(\textcolor{red}{$\uparrow$}) & 8.89(\textcolor{red}{$\uparrow$}) & 6.50(\textcolor{red}{$\uparrow$}) & 6.05(\textcolor{red}{$\uparrow$}) & 32.98(\textcolor{red}{$\uparrow$}) \\
    \cmidrule(lr){2-10}
    & Qwen2.5-VL-3B & 85.83 & 82.50 & 5.77 & 5.24 & 9.02 & 6.74 & 5.70 & 32.47 \\
    & ~+\textcolor{teal!90}{SFT}  & 86.25 & 84.58 & 2.89 & 2.22 & 4.89 & 3.52 & 2.44 & 15.96\\
    & ~+\textcolor{red!85}{RFT}  & \textbf{88.33~(\textcolor{red}{$\uparrow$2.50})} & \textbf{87.08~(\textcolor{red}{$\uparrow$4.58})} & 5.97(\textcolor{red}{$\uparrow$}) & 5.49(\textcolor{red}{$\uparrow$}) & 9.05(\textcolor{red}{$\uparrow$}) & 6.84(\textcolor{red}{$\uparrow$}) & 6.03(\textcolor{red}{$\uparrow$}) & 33.38(\textcolor{red}{$\uparrow$}) \\

    \midrule
    \multirow{10}{*}{UCF-Crime} & InternVL2.5-2B & 84.86 & 68.13 & 4.40 & 3.08 & 8.09 & 5.69 & 3.47 & 24.74 \\
    & Qwen2.5-VL-7B & 92.03 & 89.64 & 4.80 & 3.73 & 8.95 & 7.05 & 4.25 & 28.78 \\
    & InternVL 2.5 8B-MPO & 89.64 & 90.44 & 3.79 & 3.20 & 8.23 & 5.77 & 3.48 & 24.47 \\
    \cmidrule(lr){2-10}
    & Qwen2-VL-2B & 87.25 & 83.67 & 3.47 & 2.48 & 7.75 & 4.49 & 2.82 & 21.02 \\
    & ~+\textcolor{teal!90}{SFT}   & 83.67 & 86.06 & 3.61 & 2.26 & 7.30 & 4.79 & 2.70 & 20.66 \\
    & ~+\textcolor{red!85}{RFT}   & \textbf{88.45~(\textcolor{red}{$\uparrow$1.20})} & \textbf{88.05~(\textcolor{red}{$\uparrow$4.38})} & 4.04(\textcolor{red}{$\uparrow$}) & 2.75(\textcolor{red}{$\uparrow$}) & 7.72(\textcolor{blue}{$\downarrow$}) & 4.89(\textcolor{red}{$\uparrow$}) & 3.11(\textcolor{red}{$\uparrow$}) & 22.52(\textcolor{red}{$\uparrow$}) \\
    \cmidrule(lr){2-10}
    & Qwen2.5-VL-3B & 91.63 & 83.27 & 4.31 & 2.88 & 8.70 & 5.95 & 3.27 & 25.10 \\
    & ~+\textcolor{teal!90}{SFT}   & 90.84 & 90.44 & 1.80 & 1.01 & 4.15 & 2.82 & 1.11 & 10.89 \\
    & ~+\textcolor{red!85}{RFT}   & \textbf{92.03~(\textcolor{red}{$\uparrow$0.40})} & \textbf{91.63~(\textcolor{red}{$\uparrow$8.36})} & 4.42(\textcolor{red}{$\uparrow$}) & 2.98(\textcolor{red}{$\uparrow$}) & 8.71(\textcolor{red}{$\uparrow$}) & 5.98(\textcolor{red}{$\uparrow$}) & 3.39(\textcolor{red}{$\uparrow$}) & 25.49(\textcolor{red}{$\uparrow$}) \\
    \bottomrule
    \end{tabular}}
    \label{tab:msad_results}
\end{table*}

\noindent\textbf{Dataset Construction and Annotation.}
Existing video anomaly datasets typically provide only frame-level labels~\cite{sultani2018real,zhuadvancing,acsintoae2022ubnormal} or sparse descriptions~\cite{yuan2024towards,du2024exploring,du2024uncovering}, limiting their usefulness for reasoning-based tasks. To address this, we construct \textbf{VAU-Bench}, a unified benchmark built from MSAD~\cite{zhuadvancing}, UCF-Crime~\cite{sultani2018real}, and ECVA~\cite{du2024exploring}, enriched with Chain-of-Thought (CoT) annotations, including: (i) video descriptions, (ii) temporal boundaries, (iii) multiple-choice QA, and (iv) reasoning rationales. We apply a cleaning pipeline to remove corrupted or overly long videos and merge overlapping anomaly types. For UCF-Crime and ECVA, we use DeepSeek-V3~\cite{liu2024deepseek} to generate video-level summaries, QA pairs, and reasoning chains. For MSAD, CoT annotations are produced through a two-stage pipeline: we first apply InternVL-8B-MPO~\cite{wang2024enhancing} to generate initial captions and analyses, which are then verified and refined using DeepSeek-V3 to obtain more accurate QA pairs and coherent reasoning rationales. We also give further construction and annotation details in the Appendix.


\noindent\textbf{Dataset Statistics.}
Figure~\ref{fig:dataset-stat} presents an overview of VAU-Bench, the first VAU benchmark designed for Chain-of-Thought reasoning. Our dataset contains 4,602 videos covering 19 major anomaly types, with a total duration of 169.1 hours. It includes over 1.5 million words of fine-grained textual annotations, averaging 337 words per video, encompassing detailed descriptions, reasoning rationales, and multiple-choice questions. The dataset is split into 2,939 training, 734 validation, and 929 test videos. Additionally, we provide 3,700 temporal annotations to support the anomaly grounding task. Figure~\ref{fig:anomaly-type} shows the distribution of the main anomaly categories, while Figure~\ref{fig:video-length-prop} illustrates the diversity in video duration and anomaly sparsity. The evaluation protocols and metrics used for different tasks are summarized in Figure~\ref{fig:eval-criteria}, and we give more dataset statistics in the Appendix.




\noindent\textbf{Reasoning Evaluation Metric: VAU-Eval.}  
For VAU tasks, prior work has adopted BLEU and ROUGE~\cite{du2024exploring,zhang2024holmes,tang2024hawk} to evaluate semantic content.  
However, such n-gram-based metrics often fall short in capturing reasoning quality and deeper relational understanding. To better assess anomaly reasoning, we propose \textbf{VAU-Eval}, a GPT-based metric that compares model-generated descriptions and analyses with ground truth annotations. As illustrated in Figure~\ref{fig:eval-criteria}, we evaluate each response along five dimensions using DeepSeek-V3~\cite{liu2024deepseek} as the judge: classification accuracy, key concept alignment, fluency, informativeness, and factual consistency. Each dimension is scored on a 10-point scale to provide fine-grained assessment of reasoning quality.

\section{Experiment}

\noindent{\textbf{Implementation Details.}} 
Our main experiments are conducted using the Qwen2-VL-2B-Instruct~\cite{Qwen2-VL} and Qwen2.5-VL-3B-Instruct~\cite{Qwen2.5-VL} models. We apply full-parameter fine-tuning without adapters or LoRA, using 2 NVIDIA H20 GPUs for training. During the RFT training process, we adopt a structured prompting strategy that guides the model to generate intermediate reasoning and final answers in a standardized format. Specifically, each prompt instructs the model to enclose its reasoning process within \texttt{<think>...</think>} tags and its final answer within \texttt{<answer>...</answer>} tags. This format ensures consistency across different tasks. During inference, for Qwen-VL models, we sample frames at 1 FPS. For InternVL models, we uniformly sample 16 frames per video.



\subsection{Evaluation of VAU-R1}

\noindent{\textbf{Evaluation Protocol.}} We report results separately on the MSAD, ECVA, and UCF-Crime datasets rather than using a single aggregated benchmark, as these datasets differ substantially in anomaly types, video durations, and scene contexts. All evaluation metrics for our four tasks are summarized in Figure~\ref{fig:eval-criteria}.  
For the QA task, we report multiple-choice accuracy.  
Temporal anomaly grounding is evaluated using temporal mean Intersection over Union (mIoU), as well as recall at different IoU thresholds: R@0.3, R@0.5, and R@0.7.  
For anomaly reasoning, we adopt the GPT-based \textbf{VAU-Eval} introduced in Section~\ref{sec:benchmark}.  
Finally, binary and multi-class classification accuracy are used for evaluating the anomaly classification task.



\begin{table}
    \centering
    \setlength{\tabcolsep}{3pt}
    \caption{Comparison of temporal anomaly grounding performance on the three datasets. For each dataset, we present results for the base models, followed by SFT and RFT variants. \textbf{w/o think} and \textbf{w/ think} refer to the inference prompt without and with thinking, respectively. Rows highlighted in light yellow denote the results on the UCF-Crime dataset, serving as an out-of-distribution test for cross-dataset evaluation. }
    \resizebox{\textwidth}{!}{%
    \begin{tabular}{llllccc|lccc}
        \toprule
        \multirow{2.5}{*}{\textbf{Dataset}} & \multirow{2.5}{*}{\textbf{Model}} &
          & \multicolumn{4}{c|}{\textbf{w/o think}}
          & \multicolumn{4}{c}{\textbf{w/ think}} \\
        \cmidrule(lr){4-7}\cmidrule(lr){8-11}
        & & & \textbf{mIoU} & \textbf{R@0.3} & \textbf{R@0.5} & \textbf{R@0.7}
          & \textbf{mIoU} & \textbf{R@0.3} & \textbf{R@0.5} & \textbf{R@0.7} \\
        \midrule
        \multirow{5}{*}{MSAD}
            & Qwen2-VL-2B                  && 0.00  & 0.00  & 0.00  & 0.00  
                                          & 0.00  & 0.00  & 0.00  & 0.00  \\
            & Qwen2.5-VL-7B                && 45.90 & 70.83 & 45.83 & 21.67 
                                          & 17.57 & 26.67 & 11.67 & 3.33  \\
            \cmidrule(lr){2-11}
            & Qwen2.5-VL-3B                && 21.27 & 30.00 & 10.83 & 4.17  
                                          & 13.00 & 16.67 & 5.83  & 1.67  \\
            & ~+ \textcolor{teal!90}{SFT}          && 30.65 & 47.50 & 30.00 & 9.17  
                                          & 35.17 & 50.83 & 34.17 & 15.00 \\
            & ~+ \textcolor{red!85}{RFT}          && \textbf{35.77~(\textcolor{red}{$\uparrow$14.50})} & 53.33 & 34.17 & 15.83
                                          & \textbf{30.70~(\textcolor{red}{$\uparrow$17.70})} & 48.33 & 29.17 & 12.50 \\
        \midrule
        \multirow{5}{*}{ECVA}
            & Qwen2-VL-2B                  && 0.00  & 0.00  & 0.00  & 0.00  
                                          & 0.17  & 0.30  & 0.00  & 0.00  \\
            & Qwen2.5-VL-7B                && 19.85 & 25.87 & 15.17 & 9.70  
                                          & 5.71  & 7.96  & 4.73  & 2.99  \\
            \cmidrule(lr){2-11}
            & Qwen2.5-VL-3B                && 14.21 & 17.16 & 6.47  & 3.23  
                                          & 6.35  & 7.21  & 1.99  & 0.50  \\
            & ~+ \textcolor{teal!90}{SFT}          && 45.30 & 66.67 & 49.75 & 24.13 
                                          & 45.96 & 65.67 & 51.00 & 26.12 \\
            & ~+ \textcolor{red!85}{RFT}   && \textbf{35.09~(\textcolor{red}{$\uparrow$20.88})} & 49.00 & 28.86 & 19.40
                                          & \textbf{33.25~(\textcolor{red}{$\uparrow$26.90})} & 48.51 & 30.60 & 18.41 \\
        \midrule
        \multirow{5}{*}{UCF-Crime}
            & \cellcolor{yellow!10}Qwen2-VL-2B     &\cellcolor{yellow!10}& \cellcolor{yellow!10}2.74  & \cellcolor{yellow!10}4.84  & \cellcolor{yellow!10}0.00  & \cellcolor{yellow!10}0.00  
                                          & \cellcolor{yellow!10}0.12  & \cellcolor{yellow!10}0.00  & \cellcolor{yellow!10}0.00  & \cellcolor{yellow!10}0.00  \\
            & \cellcolor{yellow!10}Qwen2.5-VL-7B                &\cellcolor{yellow!10}& \cellcolor{yellow!10}22.72 & \cellcolor{yellow!10}33.87 & \cellcolor{yellow!10}16.13 & \cellcolor{yellow!10}8.06  
                                          & \cellcolor{yellow!10}4.89  & \cellcolor{yellow!10}8.06  & \cellcolor{yellow!10}1.61  & \cellcolor{yellow!10}0.00  \\
            \cmidrule(lr){2-11}
            & \cellcolor{yellow!10}Qwen2.5-VL-3B                &\cellcolor{yellow!10}& \cellcolor{yellow!10}10.91 & \cellcolor{yellow!10}15.32 & \cellcolor{yellow!10}6.45  & \cellcolor{yellow!10}3.23  
                                          & \cellcolor{yellow!10}7.68  & \cellcolor{yellow!10}10.48 & \cellcolor{yellow!10}4.84  & \cellcolor{yellow!10}1.61  \\
            & \cellcolor{yellow!10}~+ \textcolor{teal!90}{SFT}                        &\cellcolor{yellow!10}& \cellcolor{yellow!10}4.98  & \cellcolor{yellow!10}3.23  & \cellcolor{yellow!10}0.81  & \cellcolor{yellow!10}0.00  
                                          & \cellcolor{yellow!10}5.76  & \cellcolor{yellow!10}5.65  & \cellcolor{yellow!10}0.81  & \cellcolor{yellow!10}0.81  \\
            & \cellcolor{yellow!10}~+ \textcolor{red!85}{RFT}                        &\cellcolor{yellow!10}& \cellcolor{yellow!10}\textbf{16.80~(\textcolor{red}{$\uparrow$5.89})} & \cellcolor{yellow!10}23.39 & \cellcolor{yellow!10}8.06  & \cellcolor{yellow!10}4.03  
                                          & \cellcolor{yellow!10}\textbf{9.21~(\textcolor{red}{$\uparrow$1.53})}  & \cellcolor{yellow!10}9.68  & \cellcolor{yellow!10}4.03  & \cellcolor{yellow!10}1.61  \\
        \bottomrule
    \end{tabular}}%
    \label{tab:tag_results}
\end{table}

\noindent\textbf{Evaluation on QA-Guided Reasoning.}  As shown in Table~\ref{tab:msad_results}, we evaluate the reasoning capabilities of VAU-R1 on MSAD and UCF-Crime using multiple-choice QA accuracy and GPT-based VAU-Eval scores. We highlight two key observations. First, base models often perform worse when generating answers with reasoning (Acc$_\text{w/think}$) compared to without (Acc$_\text{w/o think}$) reasoning, indicating that naive Chain-of-Thought generation may introduce hallucination. In contrast, reinforcement fine-tuning (RFT) improves both QA accuracy with reasoning (\eg, +11.25 on MSAD) and overall reasoning quality. Second, RFT leads to consistent gains across five VAU-Eval dimensions—classification, demonstrating its ability to strengthen structured reasoning. For instance, on MSAD, Qwen2.5-VL-3B+RFT achieves the highest total VAU-Eval score (33.38), showing substantial improvement over its SFT counterpart. These results confirm that RFT not only enhances answer correctness but also fosters robust and generalizable multimodal reasoning under the VAU setting.






\begin{table}[t]
    \centering
    \setlength{\tabcolsep}{3pt}
    \caption{Ablation study on task co-training for anomaly classification. 
    \textbf{Bin. Acc.} denotes binary classification accuracy (normal vs. abnormal), and \textbf{Multi Acc.} denotes multi-class accuracy over 19 anomaly types plus the normal class. Results are reported with and without \textit{think} prompting. }
    \resizebox{0.75\linewidth}{!}{%
    \begin{tabular}{llcc|cc}
        \toprule
          \multirow{2.5}{*}{\textbf{Model}}  &
          & \multicolumn{2}{c|}{\textbf{w/o think}} 
          & \multicolumn{2}{c}{\textbf{w/ think}} \\
        \cmidrule(lr){3-4}\cmidrule(lr){5-6}
        && \textbf{Bin.\ Acc.} & \textbf{Multi Acc.} 
          & \textbf{Bin.\ Acc.} & \textbf{Multi Acc.} \\
        \midrule
        Baseline (Qwen2.5-VL-3B-Instruct)  
          && 62.77 & 47.96 
          & 59.33 & 39.06 \\
        ~+\textcolor{teal!90}{SFT} w/ CLS  
          && \textbf{81.12} & 29.08
          & \textbf{83.37} & 32.19 \\
        ~+\textcolor{red!85}{RFT} w/ CLS  
          && 60.30 & 46.14
          & 59.01 & 42.27 \\
        ~+\textcolor{red!85}{RFT} w/ QA 
          && 59.01 & 46.14 
          & 58.91 & 41.95 \\
        ~+\textcolor{red!85}{RFT} w/ TAG  
          && \underline{67.81} & \textbf{49.46} 
          & \underline{74.14} & \textbf{46.14} \\
        ~+\textcolor{red!85}{RFT} w/ QA-TAG  
          && 65.77 & 47.53 
          & 67.60 & 45.06 \\
        ~+\textcolor{red!85}{RFT} w/ QA-TAG-CLS  
          && 64.70 & \underline{48.61} 
          & 65.02 & \underline{45.60} \\
        \bottomrule
    \end{tabular}}%
    \label{tab:ablation}
    \vspace{-0.5cm}
\end{table}

\begin{figure}[ht]
    \centering
    \includegraphics[width=0.9\linewidth]{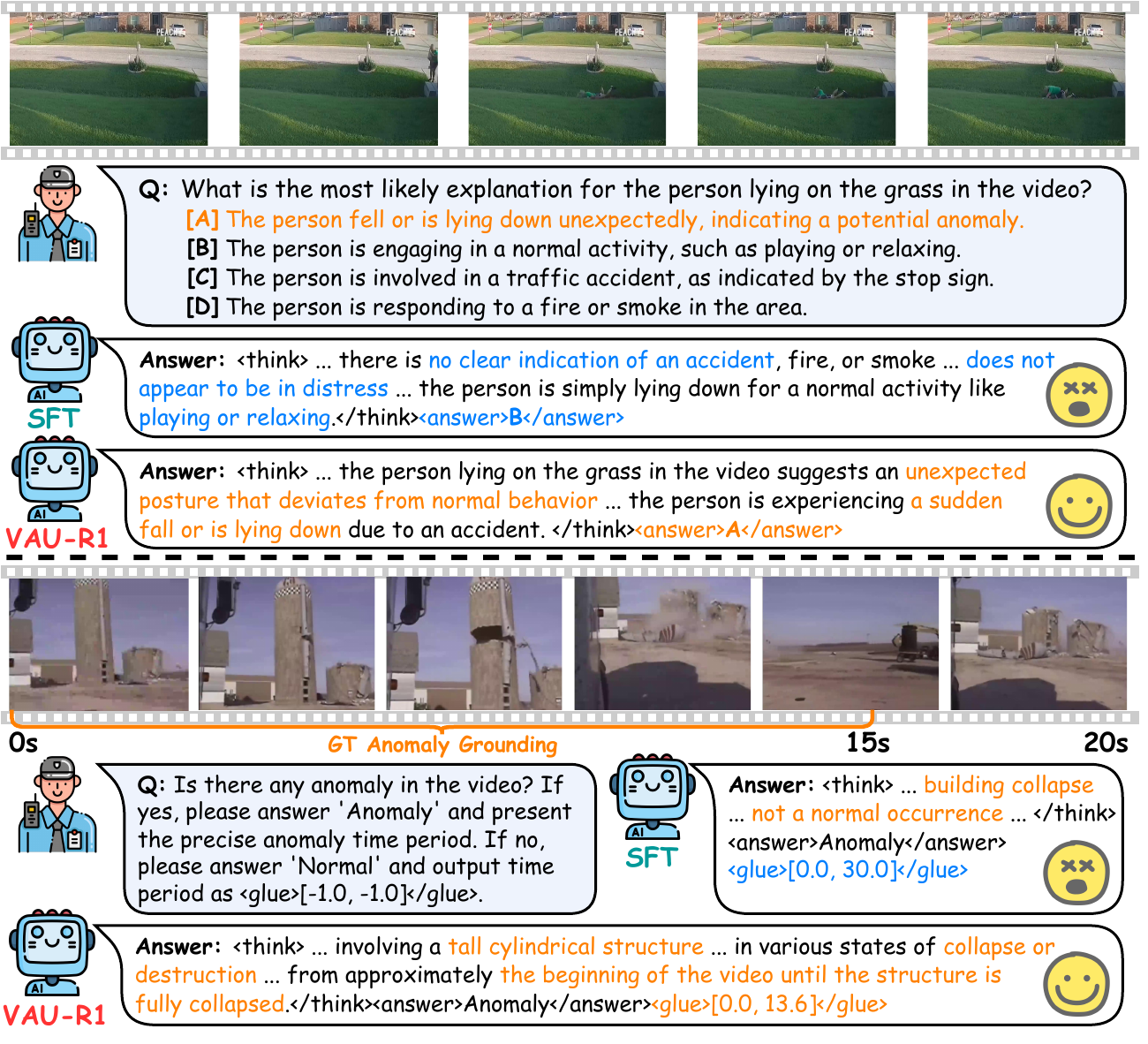}
    \caption{\textbf{Qualitative case of the QA (\textit{top}) and TAG (\textit{bottom}) task.} All ground-truths and correct answers are highlighted in orange. Both SFT and RFT perform inference using the same CoT prompt. RFT’s explicit chain-of-thought yields precise, interpretable QA choice and anomaly interval, whereas SFT’s output is less informative and tends to produce inaccurate responses.}
    \label{fig:temporal_case_study}
    \vspace{-0.5cm}
\end{figure}

\noindent\textbf{Evaluation on Temporal Anomaly Grounding.}  
As shown in Table~\ref{tab:tag_results}, we evaluate the temporal anomaly grounding performance across three datasets. Note that all models are trained only on MSAD and ECVA, while UCF-Crime serves as an out of distribution test set. We observe several key findings. First, across both inference settings (w/ and w/o think), RFT consistently outperforms the corresponding base models, demonstrating its effectiveness in improving temporal localization. Notably, the RFT-finetuned 3B model achieves higher mIoU than the larger 7B base model on ECVA. Second, similar to our observations in QA-guided reasoning, Chain-of-Thought prompting does not necessarily enhance grounding performance. In some cases, adding reasoning leads to degraded localization accuracy. Third, RFT shows significantly better generalization compared to SFT. In cross-dataset evaluation (\eg, UCF-Crime as an out-of-distribution test), SFT demonstrates limited generalization, whereas RFT maintains strong performance across unseen scenarios. While SFT occasionally outperforms RFT in isolated cases, we observe that its direct predictions are opaque and lack interpretability, often yielding repetitive, non-discriminative outputs across videos~(see Figure~\ref{fig:temporal_case_study}). These results highlight the advantages of RFT for enhancing generalization in VAU tasks.



\noindent\textbf{Ablation Study.}  
For VAU, the core objective is to make accurate high-level judgments about anomaly categories (e.g., distinguishing a \textit{fight} from a \textit{robbery}). To explore effective task formulations, we train models with different combinations of VAU tasks—multiple-choice QA, temporal anomaly grounding (TAG), and multi-class classification (CLS)—to assess their impact on reasoning. As shown in Table~\ref{tab:ablation}, RFT models trained with TAG alone achieve the highest binary accuracy (74.14) and strong multi-class performance (46.14) under the think setting, highlighting the benefit of temporal grounding for perception and category discrimination. Combining QA and TAG also improves performance but is slightly less effective than TAG alone. In contrast, SFT tends to over-predict anomalies, yielding high binary accuracy but poor multi-class results, suggesting overfitting. Overall, grounding-based tasks are more effective for anomaly classification, and jointly optimizing tasks via reinforcement learning yields complementary gains in both accuracy and reasoning.

\noindent{\textbf{Case Study.}} Figure~\ref{fig:temporal_case_study} illustrates two representative examples from the QA and TAG tasks, comparing SFT and our VAU-R1 under the same Chain-of-Thought~(CoT) prompt. In the QA example, SFT incorrectly selects a normal explanation based on surface cues, while VAU-R1 correctly infers a people-falling anomaly by identifying posture and behavioral irregularities. In the TAG example, SFT outputs a coarse anomaly span without rationale, whereas VAU-R1 localizes the anomaly more precisely (0.0--13.6s) and provides an interpretable causal chain. These cases highlight VAU-R1’s superior reasoning and interpretability in both classification and localization settings. More qualitative case studies are provided in the Appendix.

\vspace{-1em}
\subsection{Discussion}

\noindent{\textbf{RFT Enhances Generalization and Interpretability.}}  
Our experiments demonstrate that RFT consistently outperforms SFT across multiple VAU tasks, offering improved interpretability~(Table~\ref{tab:msad_results}) and better generalization~(Table~\ref{tab:tag_results}). In contrast, SFT tends to memorize task-specific patterns and suffers from poor generalization to unseen scenarios. This suggests that SFT-trained models are more prone to overfitting, especially when trained on limited or narrowly defined tasks.

\noindent{\textbf{Is Chain-of-Thought Reasoning Necessary for VAU?}}  
Our findings suggest that Chain-of-Thought (CoT) reasoning does not always lead to better performance in visual understanding tasks. However, it significantly enhances interpretability by providing structured justifications. Unlike mathematical or logical tasks, where reasoning is more deterministic, visual understanding involves inherently diverse reasoning paths. Therefore, designing simpler sub-tasks with well-defined reward signals to guide reasoning effectively remains underexplored. Directly applying complex tasks (\eg, multi-class anomaly classification) without task co-training often leads to suboptimal results (Table~\ref{tab:ablation}).


\noindent{\textbf{Rethinking Anomaly Understanding in Multimodal Contexts.}}  
VAU calls for constructing a coherent reasoning chain that bridges spatial-temporal localization and causal inference. Yet, leveraging diverse cues such as keyframes, salient objects, and even additional modalities~(\eg, audio) to support unified reasoning remains underexplored. We envision that future work could benefit from integrating these multimodal signals into a structured reasoning framework, enabling more robust and interpretable anomaly understanding. Our method and benchmark take a step in this direction by proposing a unified evaluation protocol across perception, localization, and reasoning dimensions, ultimately guiding models toward accurate and justifiable anomaly judgments.

\section{Conclusion}
We present VAU-R1, an advanced and unified Video Anomaly Understanding framework focusing on four VAU tasks: multi-choice QA, temporal grounding, anomaly reasoning, and classification. VAU-R1 leverages a multimodal large language model~(MLLM) and, notably, employs reinforcement fine-tuning to enhance anomaly reasoning and explainability via carefully designed GRPO reward functions for each task. To facilitate the training and evaluation of this framework, we also introduce VAU-Bench, the first chain-of-thought benchmark designed to train and evaluate VAU tasks at the reasoning level. The experiments on different tasks prove the strong performance of the proposed method than baselines.

{
\bibliographystyle{abbrv}
\bibliography{refs-arxiv}
}

\appendix
\newpage

\section{Further Dataset Details}
\label{sec:dataset}

\begin{figure}[ht]
    \centering
    \begin{subfigure}[ht]{0.45\linewidth}
        \centering
        \includegraphics[width=\linewidth]{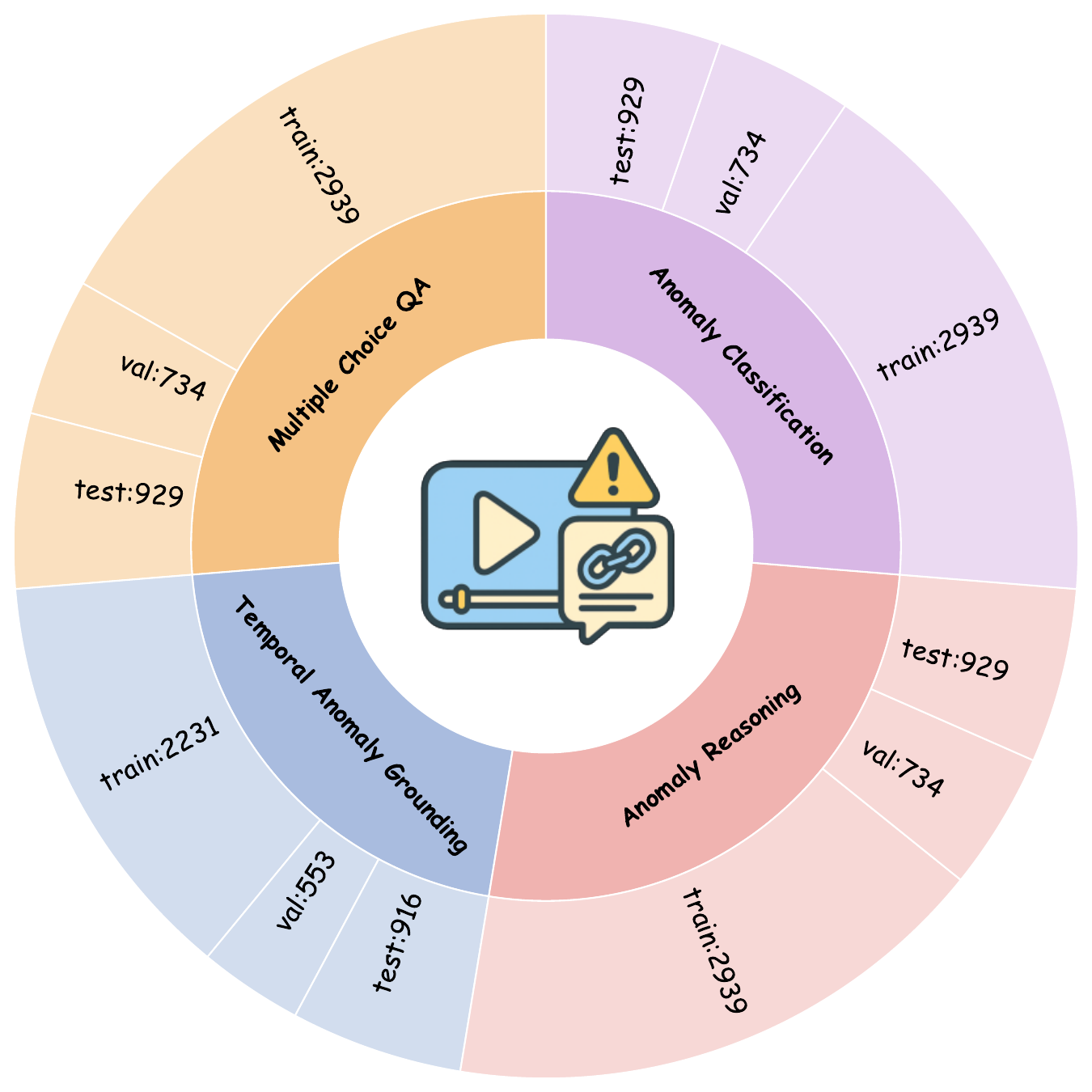}
        \caption{}
        \label{fig:train-test-split}
    \end{subfigure}
    \hspace{0.5cm}
    \begin{subfigure}[ht]{0.4\linewidth}
        \centering
        \includegraphics[width=\linewidth]{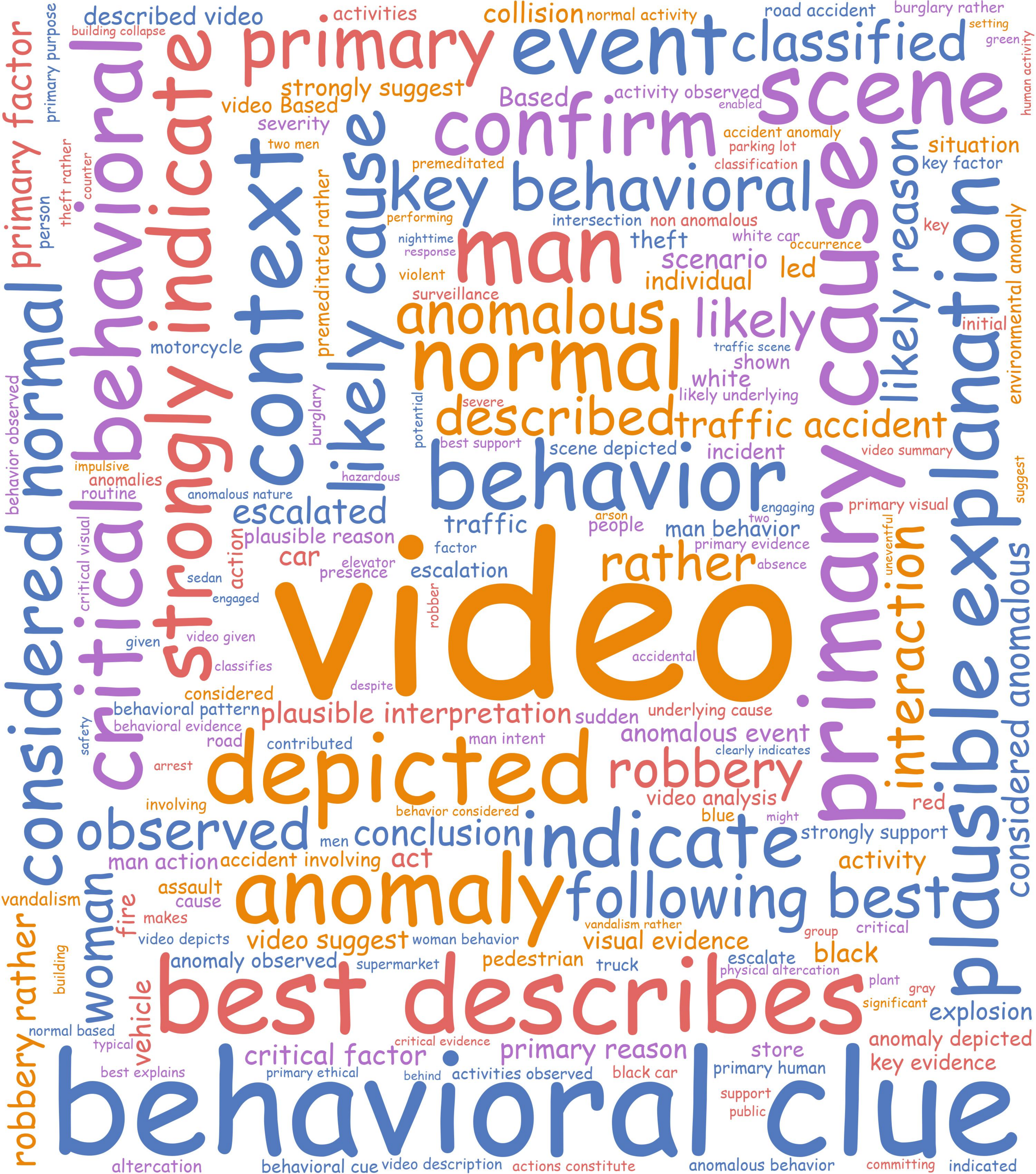}
        \caption{}
        \label{fig:word-cloud}
    \end{subfigure}
    \caption{\textbf{More dataset statistics of our VAU‐Bench.}  (a) Distribution of training, validation, and test splits across the four tasks included in VAU-Bench. (b) Word cloud visualization of frequent terms appearing in the multiple-choice questions and choices.}
    \label{fig:more-dataset-stat}
\end{figure}


\textbf{Dataset Annotation.} VAU-Bench is constructed from three datasets: UCF-Crime, ECVA, and MSAD. While UCF-Crime~\cite{sultani2018real} and ECVA~\cite{du2024exploring} provide basic scene-level descriptions, they lack the structured annotations necessary for fine-grained reasoning. To address this, we leverage DeepSeek-V3~\cite{liu2024deepseek}, a powerful large language model, to enrich the existing annotations from HIVAU-70K (which includes UCF-Crime)~\cite{zhang2024holmes} and ECVA~\cite{du2024exploring}. We use prompt-based instruction to guide the model in extracting key events, causal relationships, and anomalous behaviors, thereby producing reasoning-oriented annotations suitable for causal understanding. The detailed prompt design is provided in the blue-colored box below.

\begin{tcolorbox}[width=1.0\linewidth, colframe=black, colback=beaublue, boxsep=0mm, arc=3mm, left=1mm, right=1mm, top=2mm, bottom=2mm]
\textbf{Video Understanding Prompt.}\\
You are an expert in video understanding and reasoning. I will give you structured metadata for a surveillance or behavior-related video. Your task is twofold:

Please analyze the \textbf{entire video} description, including anomaly labels, events, and all textual summaries. Based on this, generate a comprehensive summary of what happens in the video in the following \textbf{JSON} structure:

{\small
\begin{verbatim}
{
  "judgement": "Does this video depict an anomaly? If yes, what is it called?",
  "description": "Chronological and factual summary of what happened in the video",
  "analysis": {
    "Specific Anomaly Type": "Select from the [Anomaly Type]",
    "Location": "Where the event occurs: indoor/outdoor/specific",
    "Key Evidence": "Key actions or objects that support classification",
    "Detailed Explanation": "Why these events are normal/anomalous",
    "Cause and Effect": "What led to the event and its outcome",
    "Conclusion": "Wrap-up reasoning with final conclusion about the event"
  }
}
\end{verbatim}
}

\textbf{Generating QA Pair Prompt.}\\
You are an expert in reasoning-focused QA generation for surveillance analysis videos. You will be given a structured video summary, including:
(i) A \textbf{judgement} (whether the video is anomalous or normal).
(ii) A chronological \textbf{description} of what happens in the video.
(iii) A multi-part \textbf{analysis} that breaks down the event’s anomaly type, location, key evidence, explanation, causes, and conclusion. Please generate a single multiple-choice question-answer pair in JSON format.


\end{tcolorbox}




For the MSAD~\cite{zhuadvancing} dataset, which lacks textual annotations, we design a structured Chain-of-Thought (CoT) annotation pipeline. We first use InternVL2.5-8B-MPO~\cite{wang2024enhancing} as the Vision-Language Model (VLM) to generate initial annotations that include detailed descriptions, step-by-step reasoning, and anomaly classification. To further improve the quality of these annotations, we apply DeepSeek-V3 in a secondary refinement stage, which enhances the coherence and clarity of the generated descriptions, QA pairs, and reasoning chains. The overall annotation pipeline consists of the following stages:


\begin{itemize}
    \item \textbf{Task Definition:} The VLM is instructed to act as an anomaly detector.
    \item \textbf{Video Description:} The VLM generates a detailed description of the video content.
    \item \textbf{Step-by-Step Reasoning:} The VLM performs multi-step reasoning to analyze the presence and nature of anomalies.
    \item \textbf{Verification:} Given the ground-truth anomaly type, the VLM verifies whether its prediction aligns with it. If not, it regenerates both the description and reasoning.
    \item \textbf{Key Object Summarization:} The VLM identifies key visual objects or cues relevant to the anomaly, expressed in 1–3 words.
    \item \textbf{QA Generation:} The VLM constructs multiple-choice questions by generating and shuffling plausible anomaly-related answer options.
    \item \textbf{Quality Enhancement:} We use DeepSeek-V3 to validate and refine the generated QA pairs, descriptions, and reasoning chains.
\end{itemize}

After completing the CoT annotation for the entire VAU-Bench, we perform a manual review to ensure the accuracy and consistency of all generated annotations.


\textbf{More Dataset Statistics.} Table~\ref{tab:dataset-stat} presents a detailed comparison of our \textbf{VAU-Bench} and existing video anomaly datasets. Compared to previous datasets, our benchmark offers a longer total video duration, a more diverse set of primary anomaly types (with similar categories merged), diverse multi-choice QA pairs, and richer Chain-of-Thought reasoning annotations. Figure~\ref{fig:train-test-split} shows the dataset splits across four tasks. Each task contains a balanced number of training, validation, and test samples, supporting robust evaluation. Figure~\ref{fig:word-cloud} presents a word cloud of frequent phrases extracted from the multiple-choice questions and answers in VAU-Bench. Notably, the presence of phrases such as \textit{"best describes"},  \textit{"plausible explanation"}, and \textit{"behavioral clue"} highlights the variety of question formulations, encouraging models to engage in fine-grained interpretation. In addition, keywords such as \textit{robbery}, \textit{man action}, and \textit{scene} indicate that our questions are intentionally crafted to guide models toward recognizing specific objects and anomaly types in complex real-world scenarios.

\begin{table*}
    \centering
    \setlength{\tabcolsep}{2pt}
    \caption{
    \textbf{Comparison of video anomaly detection benchmarks.} We compare VAU-Bench with existing datasets in terms of size, annotation granularity, and reasoning capabilities. VAU-Bench is the first benchmark to support structured reasoning via multiple-choice questions and Chain-of-Thought (CoT) annotations. Columns indicate whether each dataset provides QA pairs, free-text descriptions (Descrip.), anomaly judgement (Judge.), reasoning (Reason.), and full CoT rationales.
    }
    \resizebox{0.95\textwidth}{!}{\begin{tabular}{lrrrrrrrrrr}
    \toprule
    Dataset & Year & \#Videos & Total Len.  & \#Type & Annotation     & QA Pairs & Descrip. & Judge. & Reason. & CoT  \\
    \midrule
    UCSD Ped1~\cite{ucsdped2010} & 2010 & 70       & 0.1h  & 5 & Bounding-box  & \XSolidBrush & \XSolidBrush & \XSolidBrush  & \XSolidBrush & \XSolidBrush \\
    UCSD Ped2~\cite{ucsdped2010} & 2010 & 28 & 0.1h & 5 & Bounding-box & \XSolidBrush & \XSolidBrush & \XSolidBrush & \XSolidBrush & \XSolidBrush \\
    CUHK Avenue~\cite{lu2013abnormal}     & 2013 & 35       & 0.5h   & 5              & Bounding-box  & \XSolidBrush & \XSolidBrush & \XSolidBrush & \XSolidBrush & \XSolidBrush\\
    ShanghaiTech~\cite{liu2018future}  & 2017 & 437 & 3.5h & 13             & Bounding-box  & \XSolidBrush & \XSolidBrush & \XSolidBrush & \XSolidBrush & \XSolidBrush\\
    UCF‑Crime~\cite{sultani2018real}       & 2018 & 1900     & 128.0h & 13             & Frame  & \XSolidBrush & \XSolidBrush & \XSolidBrush & \XSolidBrush & \XSolidBrush\\
    Street Scene~\cite{ramachandra2020street}    & 2020 & 81  &  3.8h & 17    & Bounding-box  & \XSolidBrush & \XSolidBrush & \XSolidBrush & \XSolidBrush & \XSolidBrush\\
    IITB Corridor~\cite{Rodrigues_2020_WACV}   & 2020 & 358      &  2.0h & 10             &  Frame  & \XSolidBrush & \XSolidBrush & \XSolidBrush & \XSolidBrush  & \XSolidBrush \\
    UBNormal~\cite{acsintoae2022ubnormal}       & 2022 & 543      & 2.2h   & 22    & Frame & \XSolidBrush & \XSolidBrush & \XSolidBrush & \XSolidBrush & \XSolidBrush\\
    NWPU~\cite{cao2023new} & 2023 & 547 & 16.3h & 43 & Frame  & \XSolidBrush & \XSolidBrush & \XSolidBrush & \XSolidBrush & \XSolidBrush\\
    MSAD~\cite{zhuadvancing} & 2024 & 720  & 4.1h & 11 & Frame & \XSolidBrush & \XSolidBrush & \XSolidBrush & \XSolidBrush & \XSolidBrush\\
    \midrule
    UCA~\cite{yuan2024towards}  & 2024 & 1854 & 121.9h  & 13 & Time Duration & \XSolidBrush & \Checkmark  & \XSolidBrush & \XSolidBrush & \XSolidBrush\\
    CUVA~\cite{du2024uncovering}           & 2024 & 1000     & 32.5h  & 11     & Time Duration    & \Checkmark  & \Checkmark     & \Checkmark       & \Checkmark  & \XSolidBrush\\
    ECVA~\cite{du2024exploring}  & 2024 & 2240 & 88.2h & 21 & Time Duration   & \Checkmark  & \Checkmark  & \Checkmark       & \Checkmark & \XSolidBrush\\
    HIVAU‑70K~\cite{zhang2024holmes}       & 2025 & 5443     & NA      &  NA        & Time Duration    & \Checkmark          & \Checkmark     & \XSolidBrush   & \XSolidBrush  & \XSolidBrush \\
    \midrule
    \textbf{VAU–Bench (Ours)}   & 2025 & 4596  & \textbf{169.1h}      & 19             & Time Duration    & \Checkmark  & \Checkmark    & \Checkmark       & \Checkmark  & \Checkmark \\
    \bottomrule
        
    \end{tabular}}
    \label{tab:dataset-stat}
    \vspace{-1.5em}
\end{table*} 

\begin{figure}[t]
    \centering
    \includegraphics[width=\linewidth]{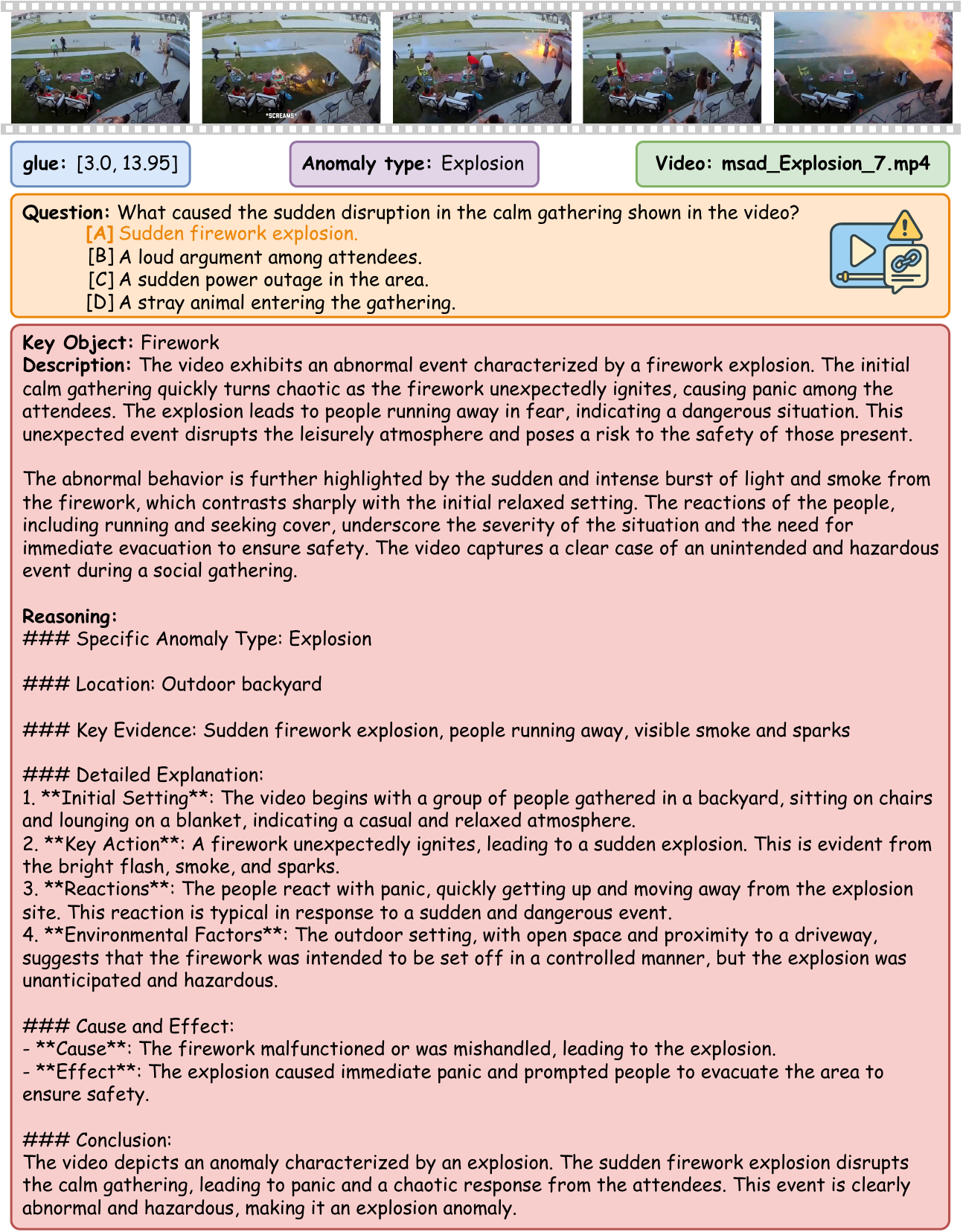}
    \caption{\textbf{Example of VAU-Bench.} An \textbf{explosion} case in an outdoor backyard, highlighting complex anomaly detection and dynamic scene understanding, labeled with a question-answer pair, key visual evidence, anomaly type, and a multi-part reasoning chain that includes location, cause and effect, and a high-level conclusion.}
    \label{fig:dataset-example-1}
\end{figure}

\begin{figure}[t]
    \centering
    \includegraphics[width=\linewidth]{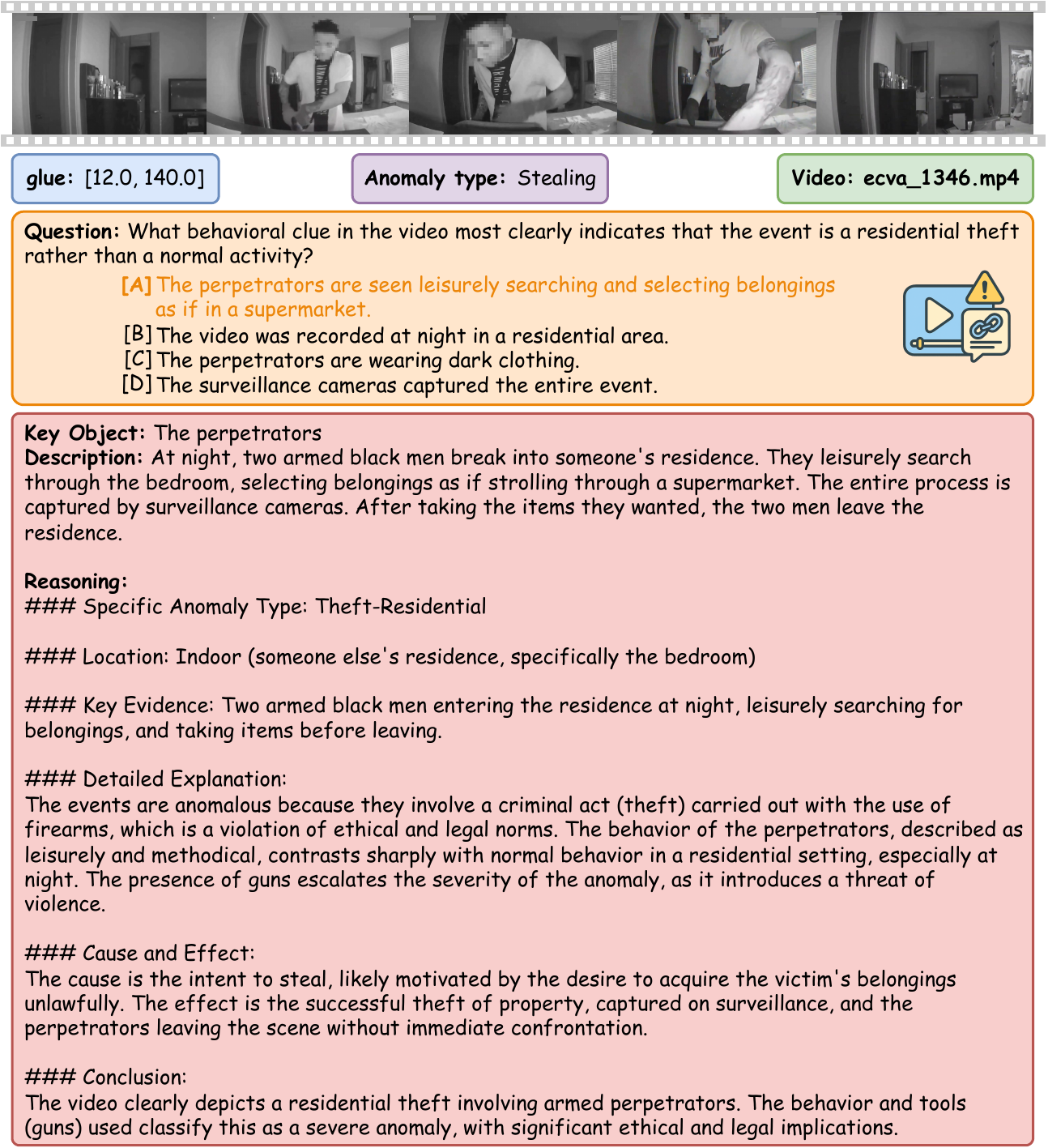}
    \caption{\textbf{Example of VAU-Bench.} A \textbf{stealing} incident, demonstrating capabilities in human activity recognition and intent analysis.}
    \label{fig:dataset-example-3}
\end{figure}

\begin{figure}[t]
    \centering
    \includegraphics[width=\linewidth]{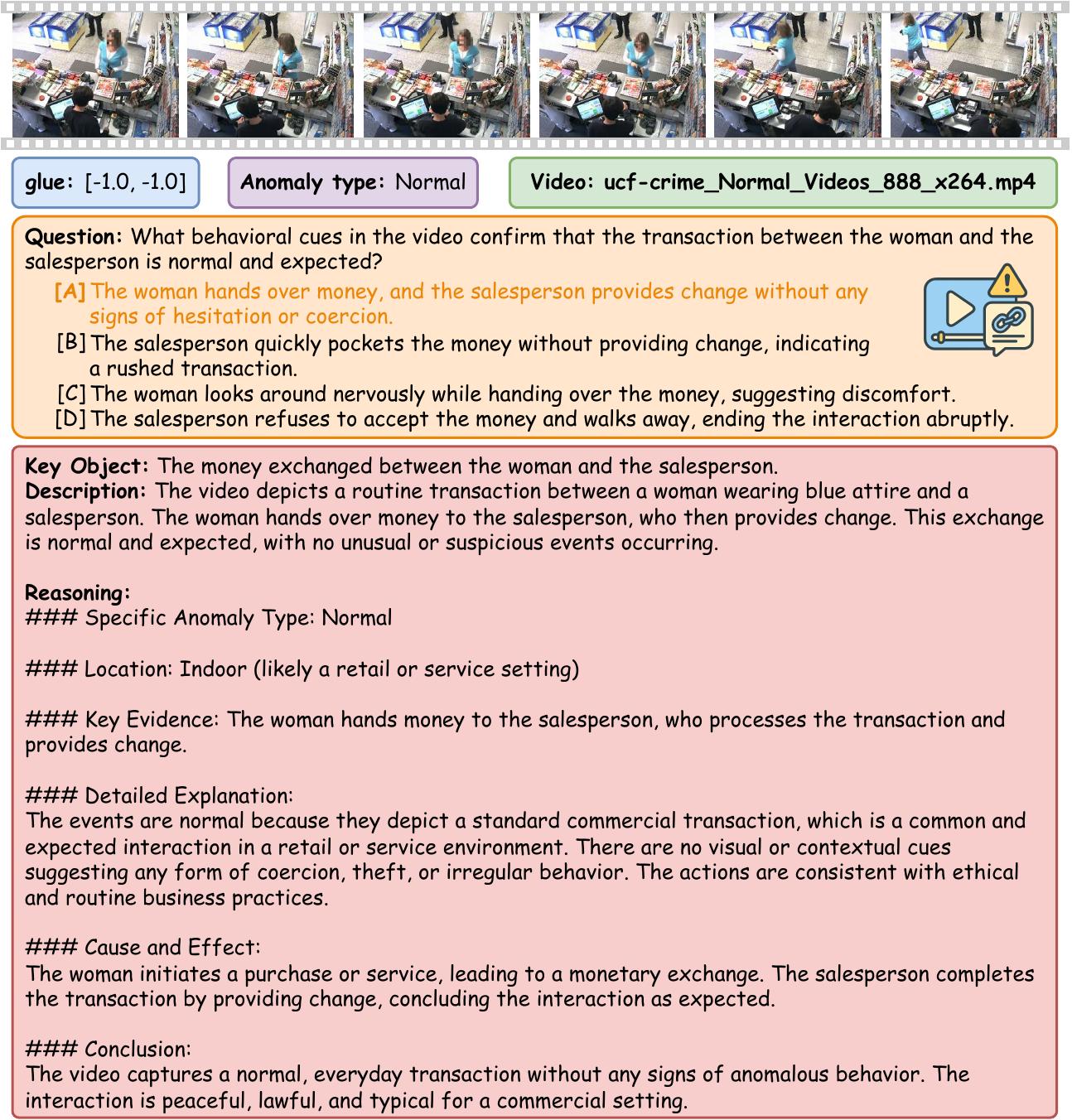}
    \caption{\textbf{Example of VAU-Bench.} A \textbf{normal} scene, used to evaluate model robustness against false positives and to enhance dataset diversity.}
    \label{fig:dataset-example-2}
\end{figure}

\textbf{Dataset Examples.} We present representative examples from our VAU-Bench, each annotated to support four core tasks of video anomaly understanding. As illustrated in Figure~\ref{fig:dataset-example-1}, each example is richly labeled with a question-answer pair, key visual evidence, anomaly type, temporal annotation, and a multi-part reasoning chain that includes location, cause and effect, and a high-level conclusion. This annotation format enables models not only to detect and classify anomalies, but also to explain them in a structured, interpretable manner. Figure~\ref{fig:dataset-example-1} and Figure~\ref{fig:dataset-example-3} show challenging anomaly scenarios, while Figure~\ref{fig:dataset-example-2} depicts a normal scene, included to test model robustness and reduce false positives. These examples demonstrate the breadth and depth of our annotations, enabling holistic evaluation across perception and reasoning dimensions.

\section{Experiment Details}
\label{sec:exp_details}
\textbf{Training Details.} We use the Adam optimizer with a learning rate of $2\times10^{-5}$. The supervised fine-tuning (SFT) stage runs for less steps (\eg 200) to avoid overfitting, while the Reinforcement Fine-Tuning (RFT) stage takes approximately 15 hours for 1.5k steps. We set the hyperparameter $\beta$ in the KL divergence term of the GRPO to 0.04, using $M=4$ candidate outputs per prompt. The maximum response length is capped at 1024 tokens.
\begin{tcolorbox}[width=1.0\linewidth, colframe=black, colback=beaublue, boxsep=0mm, arc=3mm, left=1mm, right=1mm, right=1mm, top=2mm, bottom=2mm]
\textbf{VAU-Eval Prompt.}\\
Below is a ground-truth description and analysis, followed by a model-generated description and analysis. Please evaluate the model's outputs from the following aspects:\\
1. Classification Correctness (10 pts)  \\
2. Key Object and Action Matching (10 pts)  \\
3. Fluency and Coherence (10 pts)  \\
4. Informativeness and Domain Awareness (10 pts) \\ 
5. Factual Consistency (10 pts)  
\end{tcolorbox}


\textbf{Evaluation Details for Anomaly Reasoning.} To evaluate the alignment between model-generated outputs and our annotated ground truth in video anomaly understanding, we introduce \textbf{VAU-Eval}, a GPT-based evaluation protocol. The evaluation is structured as a multi-turn interaction, where the model first generates a description of the video and then performs reasoning to determine whether the video contains an anomaly. We then use DeepSeek-V3~\cite{liu2024deepseek} to assess the similarity between the predicted answers and the ground truth across five aspects: classification correctness, key object and action matching, fluency and coherence, informativeness and domain awareness, and factual consistency. Each aspect is scored out of 10 points, yielding a total of 50 points per sample. To better reflect the model’s actual reasoning capabilities, we do not fine-tune the model on any reasoning-style description or analysis. Instead, we directly test models that are trained solely on the multiple-choice QA task, thus ensuring that their descriptive reasoning is not memorized but inferred. The detailed evaluation prompt used in this process is shown in the blue box above.



\section{Further Evaluations}
\label{sec:more_exp}



\textbf{More Evaluations.} As shown in Table~\ref{tab:ecva_results}, we conduct experiments on the ECVA dataset. Compared to UCF-Crime and MSAD, ECVA poses greater challenges across both recognition and reasoning tasks. All models consistently achieve lower  VAU-Eval reasoning scores on ECVA, indicating that its longer videos, more camera movements, viewpoint shifts and richer anomaly diversity make fine-grained understanding more difficult. While our RFT-enhanced models achieve consistent improvements in multiple-choice QA accuracy, their VAU-Eval reasoning scores does not always improve. This suggests that while RFT helps models better predict the final answer, it does not necessarily enhance the reasoning process. These findings highlight the need for more fine-grained reward signals to guide the generation of high-quality rationales in complex scenarios.


\textbf{Comparison with Prior Work.} As shown in Table~\ref{tab:holmes_results}, we evaluate HolmesVAU 2B~\cite{zhang2024holmes}, a recently released baseline for VAU, on our benchmark to assess its reasoning capability in complex scenarios. While HolmesVAU 2B achieves reasonable performance across all datasets, it consistently underperforms compared to our Qwen-based models, particularly on the challenging ECVA dataset. This performance gap is evident in both multiple-choice QA accuracy and VAU-Eval reasoning scores, indicating limitations in HolmesVAU 2B's ability to generalize to diverse and complex scenarios. In contrast, VAU-R1 demonstrates stronger alignment with human-annotated reasoning chains and greater robustness across datasets.

\begin{table*}
    \centering
    \setlength{\tabcolsep}{3pt}
    \caption{
    Comparison of performance on ECVA datasets on multiple-choice QA task and anomaly reasoning task.
    \textbf{Acc}$_{\text{w/o think}}$ and \textbf{Acc}$_{\text{w/ think}}$ refer to the multiple-choice question accuracy without and with thinking, respectively. For the anomaly reasoning task,
    \textbf{CLS}, \textbf{KM}, \textbf{FLU}, \textbf{INF}, and \textbf{FAC} represent VAU-Eval scores generated by DeepSeek-V3, measuring classification accuracy, key concept alignment, linguistic fluency, informativeness, and factual consistency, respectively. Each dimension is scored on a 10-point scale. \textbf{Total} denotes the aggregated score over five dimensions.
    }

    \resizebox{\textwidth}{!}{\begin{tabular}{llllllllll}
    \toprule
    \multirow{2.5}{*}{\textbf{Dataset}} & \multirow{2.5}{*}{\textbf{Model}} & \multicolumn{2}{c}{\textbf{QA Accuracy}} & \multicolumn{6}{c}{\textbf{VAU-Eval}} \\ 
    \cmidrule(lr){3-4} \cmidrule(lr){5-10}&     & \textbf{Acc}$_{\text{w/o think}}$ & \textbf{Acc}$_{\text{w/ think}}$ & \textbf{CLS}$\uparrow$ & \textbf{KM}$\uparrow$ & \textbf{FLU}$\uparrow$ & \textbf{INF}$\uparrow$ & \textbf{FAC}$\uparrow$ & \textbf{Total}$\uparrow$ \\ 
    \midrule
    \multirow{10}{*}{ECVA} 
    & InternVL2.5-2B & 78.84 & 58.84 & 2.86 & 2.78 & 7.57 & 4.62 & 3.03 & 20.86 \\
    & Qwen2.5-VL-7B & 83.02 & 86.98 & 3.70 & 3.67 & 8.64 & 6.40 & 4.04 & 26.45 \\
    & InternVL2.5-8B-MPO & 90.00 & 83.72 & 3.4 & 3.31 & 7.87 & 4.48 & 3.47 & 22.53 \\
    \cmidrule(lr){2-10}
    & Qwen2-VL-2B & 86.98 & 83.95 & 2.41 & 2.36 & 7.81 & 3.81 & 2.57 & 18.96 \\
    & ~+\textcolor{teal!90}{SFT}  & 84.88 & 84.65 & 2.20 & 2.12 & 7.37 & 3.99 & 2.22 & 17.90 \\
    & ~+\textcolor{red!85}{RFT}  & \textbf{90.23~(\textcolor{red}{$\uparrow$3.25})} & \textbf{84.42~(\textcolor{red}{$\uparrow$0.47})} & 2.26 & 2.28 & 7.52 & 3.70 & 2.40 & 18.16 \\
    \cmidrule(lr){2-10}
    & Qwen2.5-VL-3B & 85.58 & 75.81 & 2.21 & 2.58 & 8.33 & 5.02 & 2.75 & 20.89 \\
    & ~+\textcolor{teal!90}{SFT}  & 89.30 & 86.98 & 1.50 & 1.22 & 4.37 & 2.66 & 1.24 & 10.99 \\
    & ~+\textcolor{red!85}{RFT}  & \textbf{89.53~(\textcolor{red}{$\uparrow$3.95})} & \textbf{86.51~(\textcolor{red}{$\uparrow$10.70})} & 1.45 & 2.24 & 8.05 & 4.32 & 2.39 & 18.45 \\
    \bottomrule
    \end{tabular}}
    \label{tab:ecva_results}
\end{table*}

\begin{table*}
    \centering
    \setlength{\tabcolsep}{3pt}
    \caption{Performance of HolmesVAU 2B~\cite{zhang2024holmes} and our VAU-R1 2B on multiple-choice QA and anomaly reasoning task.}
    \resizebox{0.85\textwidth}{!}{\begin{tabular}{lllllllllll}
    \toprule
    \multirow{2.5}{*}{\textbf{Model}} && \multirow{2.5}{*}{\textbf{Dataset}} & \multicolumn{2}{c}{\textbf{QA Accuracy}} & \multicolumn{6}{c}{\textbf{VAU-Eval}} \\ 
    \cmidrule(lr){4-5} \cmidrule(lr){6-11} && & \textbf{Acc}$_{\text{w/o think}}$ & \textbf{Acc}$_{\text{w/ think}}$ & \textbf{CLS}$\uparrow$ & \textbf{KM}$\uparrow$ & \textbf{FLU}$\uparrow$ & \textbf{INF}$\uparrow$ & \textbf{FAC}$\uparrow$ & \textbf{Total}$\uparrow$ \\ 
    \midrule
    \multirow{3}{*}{HolmesVAU 2B}
    && MSAD & 85.00 & 86.25 & 3.73 & 2.72 & 6.82 & 3.55 & 3.33 & 20.15\\
    && UCF-Crime & 86.45 & 85.66 & 3.05 & 1.97 & 6.30 & 3.08 & 2.39 & 16.79\\
    && ECVA & 70.47 & 70.70 & 2.54 & 1.71 & 6.26 & 2.78 & 2.30 & 15.59\\
    \midrule
    \multirow{3}{*}{VAU-R1 2B}
    && MSAD & 82.92~(\textcolor{blue}{$\downarrow$2.08}) & 83.75~(\textcolor{blue}{$\downarrow$2.50}) & 6.05 & 5.49 & 8.89 & 6.50 & 6.05 & 32.98 \\
    && UCF-Crime & \textbf{88.45~(\textcolor{red}{$\uparrow$2.00})} & \textbf{88.05~(\textcolor{red}{$\uparrow$2.39})} & 4.04 & 2.75 & 7.72 & 4.89 & 3.11 & 22.52\\
    && ECVA & \textbf{90.23~(\textcolor{red}{$\uparrow$19.76})} & \textbf{84.42~(\textcolor{red}{$\uparrow$13.72})} & 2.26 & 2.28 & 7.52 & 3.70 & 2.40 & 18.16 \\
    \bottomrule
    \end{tabular}}
    \label{tab:holmes_results}
\end{table*}

\textbf{Classification Results.} Table~\ref{tab:cls_results} presents the binary and multi-class anomaly classification accuracy on three datasets: MSAD, UCF-Crime, and ECVA. We directly apply the RFT strategy to train a multi-class anomaly classification task, which includes 19 different anomaly types as well as the normal class. However, directly training the complex multi-class task with RFT degrades performance, suggesting it is more effective to decompose the task into simpler sub-tasks with structured rewards to better guide learning. We compare multiple models under two settings: \textbf{w/o think} and \textbf{w/ think}. We observe that, for the relatively challenging multi-class anomaly task, incorporating an explicit “think” reasoning step improves the model’s classification accuracy.


\begin{table}
    \centering
    \setlength{\tabcolsep}{3pt}
    \caption{Comparison of anomaly classification accuracy on three datasets. 
    \textbf{Bin. Acc.} denotes binary classification accuracy (normal vs. abnormal), and \textbf{Multi Acc.} denotes multi-class accuracy over 19 anomaly types and the normal class. Results are reported with and without \textit{think} prompting.}
    \resizebox{0.85\textwidth}{!}{%
    \begin{tabular}{ll|cc|cc}
        \toprule
        \multirow{2}{*}{\textbf{Dataset}} & \multirow{2}{*}{\textbf{Model}}
          & \multicolumn{2}{c|}{\textbf{w/o think}}
          & \multicolumn{2}{c}{\textbf{w/ think}} \\
        \cmidrule(lr){3-4}\cmidrule(lr){5-6}
        & & \textbf{Bin. Acc.} & \textbf{Multi Acc.}
          & \textbf{Bin. Acc.} & \textbf{Multi Acc.} \\
        \midrule
        \multirow{5}{*}{MSAD}
            & Qwen2-VL-2B-Instruct        & 75.00 & 62.50 & 60.42 & 52.92 \\
            & Qwen2.5-VL-7B-Instruct      & 90.00 & 70.00 & 75.00 & 66.67 \\
            \cmidrule(lr){2-6}
            & Qwen2.5-VL-3B-Instruct      & 79.17 & 69.58 & 73.33 & 56.67 \\
            & ~+ \textcolor{teal!90}{SFT} & 70.83 & 28.75 & 74.58 & 33.33 \\
            & ~+ \textcolor{red!85}{RFT}  
                & \textbf{82.08~(\textcolor{red}{$\uparrow$2.91})} 
                & \textbf{71.25~(\textcolor{red}{$\uparrow$1.67})} 
                & \textbf{74.58~(\textcolor{red}{$\uparrow$1.25})}
                & \textbf{60.83~(\textcolor{red}{$\uparrow$4.16})} \\
        \midrule
        \multirow{5}{*}{UCF-Crime}
            & Qwen2-VL-2B-Instruct        & 60.56 & 53.78 & 60.16 & 51.79 \\
            & Qwen2.5-VL-7B-Instruct      & 86.85 & 62.15 & 70.12 & 61.35 \\
            \cmidrule(lr){2-6}
            & Qwen2.5-VL-3B-Instruct      & 64.54 & 58.57 & 62.55 & 52.19 \\
            & ~+ \textcolor{teal!90}{SFT} & 64.14 & 28.69 & 69.32 & 37.05 \\
            & ~+ \textcolor{red!85}{RFT}  
                & 62.55~(\textcolor{blue}{$\downarrow$1.99})
                & 57.77~(\textcolor{blue}{$\downarrow$0.80}) 
                & 62.15~(\textcolor{blue}{$\downarrow$0.40})
                & \textbf{56.97~(\textcolor{red}{$\uparrow$4.78})} \\
        \midrule
        \multirow{5}{*}{ECVA}
            & Qwen2-VL-2B-Instruct        & 41.95 & 24.72 & 32.88 & 19.05 \\
            & Qwen2.5-VL-7B-Instruct      & 64.85 & 32.88 & 43.54 & 23.81 \\
            \cmidrule(lr){2-6}
            & Qwen2.5-VL-3B-Instruct      & 52.83 & 30.16 & 49.89 & 22.00 \\
            & ~+ \textcolor{teal!90}{SFT} & 96.37 & 29.48 & 96.15 & 28.80 \\
            & ~+ \textcolor{red!85}{RFT}  
                & 49.66~(\textcolor{blue}{$\downarrow$3.17})
                & \textbf{30.61~(\textcolor{red}{$\uparrow$0.45})} 
                & \textbf{55.78~(\textcolor{red}{$\uparrow$5.89})}
                & \textbf{31.07~(\textcolor{red}{$\uparrow$9.07})} \\
        \bottomrule
    \end{tabular}}%
    
    \label{tab:cls_results}
\end{table}

\begin{table}[t]
\centering
\caption{Comparison of temporal localization performance (mIoU) across different methods on UCF-Crime dataset.}
\resizebox{0.61\textwidth}{!}{
\label{tab:tiou}
\begin{tabular}{lllr}
    \toprule
    \textbf{Category} & \textbf{Method} & \textbf{Feature} & \textbf{mIoU} \\
    \midrule
    \multirow{8}{*}{Traditional} 
    & Two-stream~\cite{simonyan2014two} & Two-Stream & 2.20  \\
    & TSN~\cite{wang2016temporal}   & TSN & 2.60  \\
    & C3D~\cite{tran2015learning}   & C3D & 7.20  \\
    & T-C3D~\cite{liu2018t}    & C3D & 10.20 \\
    & ARTNet~\cite{wang2018appearance}  & ARTNets & 11.40 \\
    & 3DResNet~\cite{hara2018can}  & I3D-ResNet & 10.30 \\
    & NLN~\cite{wang2018non}       & I3D-ResNet & 12.20 \\
    & Liu \etal~\cite{liu2019exploring} & I3D-ResNet & 16.40 \\
    \midrule
    \multirow{2}{*}{Multi-modal} 
    & VADClip~\cite{wu2024vadclip}             & CLIP & 22.05 \\
    & STPrompt~\cite{wu2024weakly}  & CLIP & 23.90 \\
    \midrule
    \multirow{3}{*}{MLLMs} 
    & Qwen2.5-VL-3B        & ViT & 10.91 \\
    & Qwen2.5-VL-3B~+~\textcolor{red!85}{RFT}  & ViT & 16.80 \\
    & Qwen2.5-VL-7B  & ViT & 22.72 \\
    \bottomrule
\end{tabular}
}
\end{table}

\textbf{Temporal Localization Performance.}
Table~\ref{tab:tiou} summarizes the temporal localization (mIoU) performance of representative methods, categorized into traditional models, multi-modal approaches, and MLLMs. As expected, early appearance-based methods (\eg, Two-stream~\cite{simonyan2014two}, TSN~\cite{wang2016temporal}, C3D~\cite{tran2015learning}) achieve limited performance. Incorporating spatio-temporal modeling via 3D convolutions (T-C3D~\cite{liu2018t}, ARTNet~\cite{wang2018appearance}, 3DResNet~\cite{hara2018can}) brings moderate improvements, with Liu~\etal~\cite{liu2019exploring} reaching a mIoU of 16.40. More recent multi-modal approaches, such as VADClip~\cite{wu2024vadclip} and STPrompt~\cite{wu2024weakly}, achieve significantly better performance, with STPrompt reaching 23.90 mIoU. Our MLLM-based methods show promising yet limited temporal grounding capabilities. While Qwen2.5-VL-3B achieves only 10.91 mIoU, reinforcement tuning (+RFT) boosts performance to 16.80, indicating that structured reward learning helps align model outputs with temporal structures. However, even with RFT, MLLMs still underperform compared to specialized temporal models, suggesting that current architectures may lack explicit temporal reasoning modules required for fine-grained localization.

\begin{figure}[t]
    \centering
    \includegraphics[width=\linewidth]{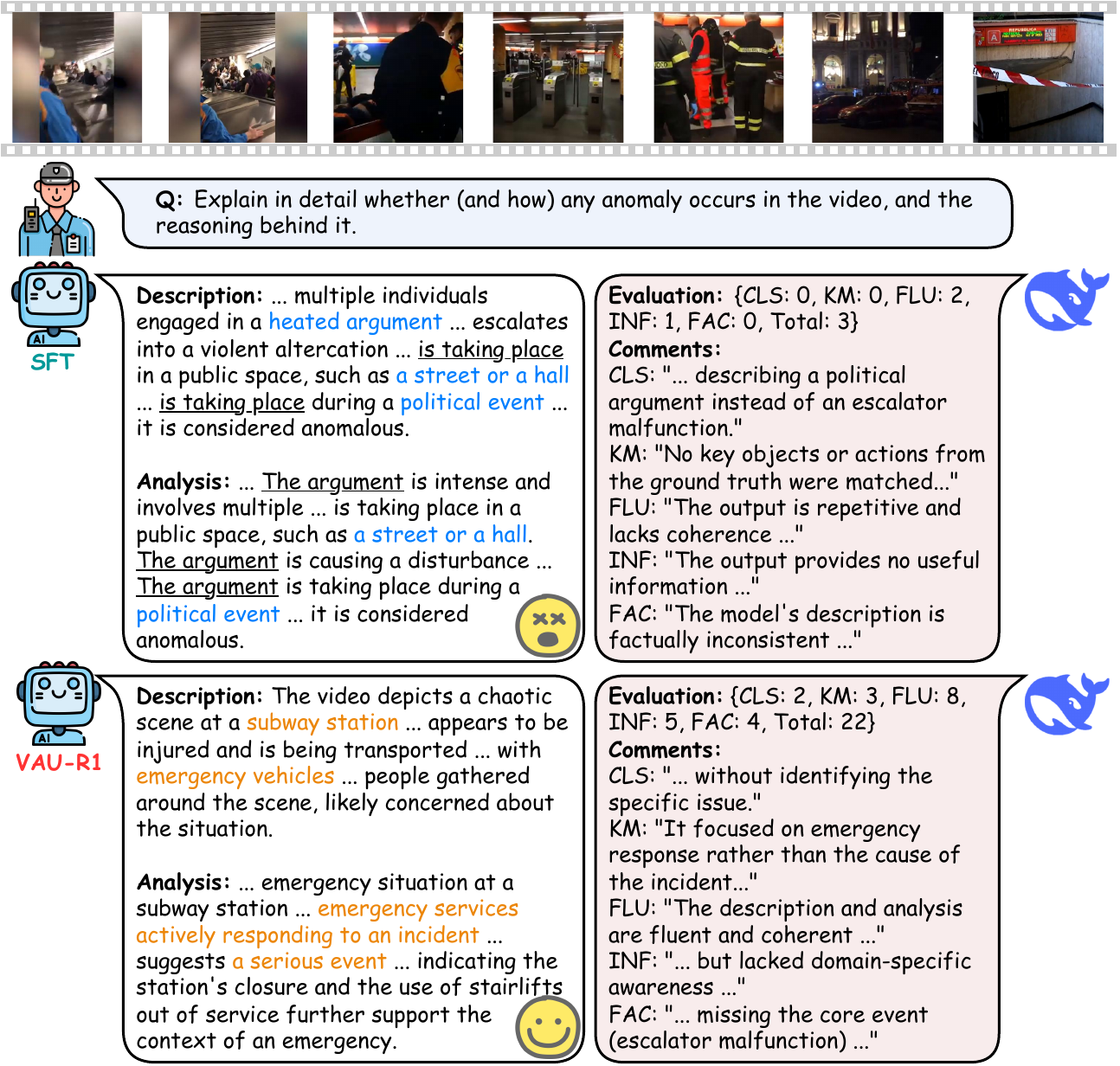}
    \caption{\textbf{Qualitative case of the Anomaly Reasoning task.} All correct description and analysis are highlighted in orange. The evaluation results are presented on the right of the answer respectively. Both SFT and VAU-R1 perform inference using the same CoT prompt. VAU-R1’s output correctly identifies the anomaly with high fluency but lacks reasoning for the core event, whereas SFT’s output is inaccurate and tends to produce repetitive responses.}
    \label{fig:description_case_study}
\end{figure}

\textbf{Case Study on Anomaly Reasoning.} Figure~\ref{fig:description_case_study} presents a qualitative comparison between outputs generated by SFT and our proposed VAU-R1 model on anomaly reasoning task. Both models are evaluated using the same Chain-of-Thought (CoT) prompt and scored based on five criteria: classification correctness (CLS), key object matching (KM), fluency (FLU), informativeness (INF), and factual consistency (FAC). The SFT output incorrectly identifies the anomaly as a political argument, which does not match the core issue (an escalator malfunction). It also fails to mention any key visual evidence or relevant location. In contrast, VAU-R1 produces a more contextually appropriate response, identifying an emergency situation at a subway station involving injured individuals and emergency vehicles. While the response focuses on surface-level emergency context rather than the root cause, it demonstrates greater fluency and relevance. The evaluation assigns a higher total score of 22, with solid performance across all dimensions, particularly in fluency and informativeness.




\section{Limitation and Future Work}
\label{sec:limitation}

One limitation of this work is its focus on a constrained set of tasks, namely multiple-choice question answering, temporal grounding, anomaly reasoning, and anomaly classification. While these tasks form a strong foundation for video anomaly understanding, there remains substantial room for extension. Future work could incorporate additional tasks such as spatial localization of key objects, which would enable more fine-grained event understanding. Moreover, introducing additional modalities~(\eg, audio) may provide complementary cues that enhance both the robustness and contextual depth of anomaly reasoning.

\section{Potential Societal Impact}
\label{sec:social_impact}
We propose a new method and benchmark for video anomaly understanding. Accurate and interpretable anomaly understanding systems can contribute to a wide range of safety-critical applications, such as disaster early warning, fire prevention, fall detection, and public safety monitoring. By enabling models to reason about abnormal events, our approach can assist first responders in identifying urgent situations earlier and more reliably. 

However, this research inevitably involves scenarios that depict violent or chaotic abnormal behaviors. We strictly follow established ethical guidelines throughout our study. The datasets used in this study are publicly available and have been processed in accordance with the guidelines provided by their original publishers. We strictly adhere to these terms of use and employ the data solely for academic research purposes. To ensure privacy protection, the datasets include safeguards such as reduced video resolution and facial blurring, effectively preventing the identification of individuals. Looking ahead, we plan to explore anomaly understanding methods that incorporate privacy preservation as a core design principle.

\newpage

\clearpage



\newpage

\end{document}